\global\def\draftcontrol{0}

   \def\versionno{RTBM v3}

\catcode`\@=11

\expandafter\ifx\csname draftcontrol\endcsname\relax\global\def\draftcontrol{0} 
\fi 

{\count255=\time\divide\count255 by 60 
\xdef\hourmin{\number\count255} 
\multiply\count255 by-60\advance\count255 by\time 
\xdef\hourmin{\hourmin:\ifnum\count255<10 0\fi\the\count255}} 
\def\draftdate{\number\month/\number\day/\number\year\ \ \ \hourmin } 

\newcommand\makepapertitle{\par

  \begingroup 
    \renewcommand\thefootnote{\@fnsymbol\c@footnote}%
    \def\@makefnmark{\rlap{\@textsuperscript{\normalfont\@thefnmark}}}%
    \long\def\@makefntext##1{\parindent 1em\noindent 
            \hb@xt@1.8em{%
                \hss\@textsuperscript{\normalfont\@thefnmark}}##1}%
     \newpage 
     \global\@topnum\z@   
     \@makepapertitle 
     \thispagestyle{empty}\@thanks 
  \endgroup 
  \setcounter{footnote}{0}%
  \global\let\thanks\relax 
  \global\let\makepapertitle\relax 
  \global\let\@makepapertitle\relax 
  \global\let\@thanks\@empty 
  \global\let\@author\@empty 
  \global\let\@date\@empty 
  \global\let\@title\@empty 
  \global\let\title\relax 
  \global\let\author\relax 
  \global\let\date\relax 
  \global\let\and\relax 
  \def\version{\let\version\@version\@gobble} 
} 
\def\@makepapertitle{%
  \newpage 
   \ifnum\draftcontrol=1 {} 
   \version\versionno 
   \vskip 5.5em%
   \else 
   \hfill\hbox to 3cm {\parbox{4.5cm}{\@pubnum}\hss}%
   \vskip 6.5em%
   \fi 
   \begin{center}%
   \let \footnote \thanks 
      {\hskip -0\textwidth \hbox to 1\textwidth%
        {\centerline{\Large\bf{\noindent\@title}}}}%
     \vskip 2em%
     {\normalsize
       \lineskip .5em%
       \begin{tabular}[t]{c}%
         \@author 
       \end{tabular}\par}%
     \vskip 1.5em%
     {\@bstract}%
     \end{center}%
     \vfill
     \@date%
     \vskip 1.5em%
   \par 
} 

\gdef\@pubnum{} 
\def\pubnum#1{%
  \gdef\@pubnum{#1}} 

\gdef\@bstract{} 
\def\Abstract#1{%
  \gdef\@bstract{%
   \parbox{\textwidth-0pc}{%
   \centerline{\bf Abstract}\penalty1000 
   \noindent
   \renewcommand\baselinestretch{1.0} 
   {#1}}} 
} 

\gdef\@email{}
\def\email#1{%
   \gdef\@email{%
   Email: {\tt #1}}
}

\def\ps@paper{\let\@mkboth\@gobbletwo%
     \ifnum\draftcontrol=1 
        \def\@oddfoot{\hbox to \textwidth{\tiny \versionno \hfil\tiny\draftdate}%
        \hskip -\textwidth \hbox to \textwidth{\hfil\rm\thepage\hfil}}%
     \else\def\@oddfoot{\hbox to \textwidth{\hfil\rm\thepage\hfil}} 
     \fi 
     \let\@evenfoot\@oddfoot 
} 

\def\body{\clearpage 
          \pagestyle{paper} 
        } 
\newenvironment{acknowledgments}{%
\vskip 3.25ex 
\addcontentsline{toc}{section}{Acknowledgments}
\noindent {\bf Acknowledgments} 
} 

\def\@version#1{\ifnum\draftcontrol=1 
\typeout{}\typeout{#1}\typeout{} 
\vskip3mm\centerline{\hbox{\fbox{\normalsize{\tt DRAFT -- #1 -- } 
                   {\draftdate}}}}\vskip3mm 
\fi} 
\let\version\@version 
\long\def\eqlabel#1{\ifnum\draftcontrol=1 
                    \tag@false  
                    \tag*{(\theequation) \hbox to -0.2cm{\hspace{0cm}\small{#1}\hss}} 
                    \refstepcounter{equation}  
                    \edef\@currentlabel{\theequation} 
                    \ltx@label{#1}          
                    \else 
                    \label{#1} 
                    \fi 
                    } 
\let\st@bibitem\@bibitem 
\let\st@lbibitem\@lbibitem 
\ifnum\draftcontrol=1 
  \def\@bibitem#1{%
    \st@bibitem{#1}\a@@label{#1}\ignorespaces} 
  \def\@lbibitem[#1]#2{%
    \st@lbibitem[#1]{#2}\a@@label{#2}\ignorespaces} 
  \def\a@@label#1{%
    \gdef\a@lab{\smash{\normalfont\small#1}} 
    \ifvmode 
      \if@inlabel 
        \global\setbox\@labels\hbox{%
          \llap{\a@lab\let\a@lab\relax 
                \kern\@totalleftmargin\kern\marginparsep}%
          \box\@labels}%
      \fi 
    \fi} 
\fi 

\documentclass[12pt,letterpaper]{article} 

\usepackage{amsmath,bm,amsfonts,amssymb,array,calc,amsthm,rotating,amscd}
\usepackage{epsfig,psfrag} 
\usepackage{rotating}
\usepackage{amscd}
\usepackage{graphicx}
\usepackage{color}
\usepackage[colorlinks=false]{hyperref}

\renewcommand\baselinestretch{1.25} 
\renewcommand\section{\@startsection {section}{1}{\z@}%
                                   {-3.5ex \@plus -1ex \@minus -.2ex}%
                                   {2.3ex \@plus.2ex}%
                                   {\normalfont\large\bfseries}} 
\renewcommand\subsection{\@startsection{subsection}{2}{\z@}%
                                   {-3.25ex\@plus -1ex \@minus -.2ex}%
                                   {1.5ex \@plus .2ex}%
                                   {\normalfont\normalsize\bfseries}} 
\renewcommand\subsubsection{\@startsection{subsubsection}{3}{\z@}%
                                   {-3.25ex\@plus -1ex \@minus -.2ex}%
                                   {1.5ex \@plus .2ex}%
                                   {\normalfont\normalsize\it}} 
\renewcommand\paragraph{\@startsection{paragraph}{4}{\z@}%
                                   {-1.75ex\@plus -1ex \@minus -.2ex}%
                                   {1ex \@plus .2ex}%
                                   {\normalfont\normalsize\bf}} 
\renewcommand\subparagraph{\@startsection{subparagraph}{5}{\z@}%
                                   {-1.25ex\@plus -0ex \@minus -.2ex}%
                                   {-2ex \@plus .2ex}%
                                   {\normalfont\normalsize\it}}

\numberwithin{equation}{section}

\long\def\@makecaption#1#2{%
  \vskip\abovecaptionskip
  \sbox\@tempboxa{{\bf #1:} #2}%
  \ifdim \wd\@tempboxa >\hsize
    {\small\bf #1:} {\small #2}\par
  \else
    \global \@minipagefalse
    \hb@xt@\hsize{\hfil\box\@tempboxa\hfil}%
  \fi
  \vskip\belowcaptionskip}

\renewcommand*\l@section[2]{%
  \ifnum \c@tocdepth >\z@
    \addpenalty\@secpenalty
    \addvspace{.5em \@plus\p@}%
    \setlength\@tempdima{1.5em}%
    \begingroup
      \parindent \z@ \rightskip \@pnumwidth
      \parfillskip -\@pnumwidth
      \leavevmode \bfseries
      \advance\leftskip\@tempdima
      \hskip -\leftskip
      #1\nobreak\hfil \nobreak\hb@xt@\@pnumwidth{\hss #2}\par
    \endgroup
  \fi}
\renewcommand*\l@subsection{\addvspace{.0em \@plus\p@}\@dottedtocline{2}{1.5em}{2.3em}}
\renewcommand*\l@subsubsection{\addvspace{-.2em \@plus\p@}\@dottedtocline{3}{3.8em}{3.2em}}

\def\arxiv#1#2{\href{http://xxx.arxiv.org/abs/#1}{{arXiv:#1 [#2]}}}

\definecolor{refcol}{rgb}{0.2,0.2,0.8}
\definecolor{eqcol}{rgb}{.6,0,0}
\definecolor{purple}{cmyk}{0,1,0,0}

\gdef\@citecolor{refcol}
\gdef\@linkcolor{eqcol}
\def\colorlinkspurple{\gdef\@urlcolor{purple}}
\def\colorlinksblue{\gdef\@urlcolor{blue}}
\def\colorlinksred{\gdef\@urlcolor{red}}

\def\ie{{\it i.e.}}

\def\cf{{\it cf.}}

\def\revise#1       {\raisebox{-0em}{\rule{3pt}{1em}}%
                     \marginpar{\raisebox{.5em}{\vrule width3pt\ 
                     \vrule width0pt height 0pt depth0.5em 
                     \hbox to 0cm{\hspace{0cm}{%
                     \parbox[t]{4em}{\raggedright\footnotesize{#1}}}\hss}}}}

\def\ii           {{\mathrm i}}

\def\sqr#1#2{{\vcenter{\vbox{\hrule height.#2pt   
 \hbox{\vrule width.#2pt height#1pt \kern#1pt 
 \vrule width.#2pt}\hrule height.#2pt}}}}

\newcommand{\Z}{\mathbb Z}

\newcommand{\beq}{\begin{equation}}
\newcommand{\eq}{\end{equation}}
\newcommand{\req}[1]{(\ref{#1})}

\catcode`\@=12

\begin{document} 


\title{Riemann-Theta Boltzmann Machine}

\pubnum{CERN-TH-2017-275}
\date{December 2017}

\author{Daniel Krefl$^{a,1}$, Stefano Carrazza$^{a,2}$, Babak Haghighat$^{b,3}$, and Jens Kahlen$^{c,4}$
 \\[0.2cm]
\it  $^a$ Theoretical Physics Department, CERN, Geneva 23, CH-1211 Switzerland\\
\it $^b$ Yau Mathematical Sciences Center, Tsinghua University, Beijing, 100084, China\\
\it $^c$ Independent Researcher, Am Dachsbau 8, 53757 Sankt Augustin, Germany\\
\it \scriptsize{$^1$ daniel.krefl@cern.ch, $^2$ stefano.carrazza@cern.ch, $^3$ babak.haghighat@gmail.com, $^4$ jens.kahlen@gmail.com}
}

\Abstract{
A general Boltzmann machine with continuous visible and discrete integer valued hidden states is introduced. Under mild assumptions about the connection matrices, the probability density function of the visible units can be solved for analytically, yielding a novel parametric density function involving a ratio of Riemann-Theta functions. The conditional expectation of a hidden state for given visible states can also be calculated analytically, yielding a derivative of the logarithmic Riemann-Theta function. The conditional expectation can be used as activation function in a feedforward neural network, thereby increasing the modelling capacity of the network. Both the Boltzmann machine and the derived feedforward neural network can be successfully trained via standard gradient- and non-gradient-based optimization techniques.
}

\makepapertitle

\body

\version\versionno

\vskip 1em



\section{Introduction}
In this work we introduce a new variant of the Boltzmann machine, a type of stochastic recurrent neural network first proposed by Hinton and Sejnowski \cite{HS1983}. Restricted versions of Boltzmann machines have been successfully used in many applications, for example dimensional reduction \cite{HS2006}, generative pretraining \cite{HDYDMJS2012}, learning features of images \cite{K2009}, and as building blocks for hierarchical models like Deep Believe Networks, \cf, \cite{S2009} and references therein.

Unlike the original Boltzmann machine, the partition function of our new variant, and thus the visible units' probability density function, can be solved for analytically.\footnote{For clarification, we understand under {\it analytically} that we can write down a closed form functional representation. The explicit evaluation of the functional representation requires a numerical approximation. Also note that the statement holds for the fully connected case. For analytic approaches to the restricted Boltzmann machine, see \cite{B2017,B2018}.}  Hence, we do not need to invoke the usual learning algorithms for (restricted) Boltzmann machines such as Contrastive Divergence \cite{HOT2006}. The resulting probability density function we obtain constitutes a new class of parametric probability densities, generalizing the multi-variate Gaussian distribution in a highly non-trivial way. We have to make certain assumptions about the connection matrices of our new variant of the Boltzmann machine, but they are rather mild: Namely, the self-connections in both the visible and hidden sector have to be real, symmetric and positive definite. Note that we explicitly allow for self-couplings of the network nodes. The connection matrix which couples the two sectors needs to be either purely real or imaginary. Furthermore, in the real case the overall connection matrix needs to be positive definite as well. The setup is illustrated in figure \ref{RTBMfig}. 

If we take the visible and hidden sector states to be continuous in $\mathbb R$ (we will refer to this as a continuous Boltzmann machine), it is easy to show that the corresponding probability density is simply the multi-variate normal distribution, \cf, appendix \ref{CBM}. In contrast, our new version of the Boltzmann machine has a continuous visible sector, but the hidden sector states are restricted to take discrete integer values. One may see this as a form of quantization of the continuous Boltzmann machine. The case of a finite number of discrete hidden states has been considered in \cite{WRH2004}. In the setup we discuss in this work, each hidden node possesses an infinite amount of different states. The set of states of a single node is $\mathbb Z$, and therefore the hidden state space is $\Z^{N_h}$, where $N_{h}$ is the number of hidden units. This is a generalized version of the Gaussian-Bernoulli Boltzmann machine, \cf, \cite{WRH2004,HS2006,B2012,H2012}, which has continuous visible units and a binary hidden sector. 

We will refer to our new variant of the Boltzmann machine as the Riemann-Theta Boltzmann machine (or RTBM for short). As derived later in the paper, the closed form solution of the probability density function of the visible units reads
\beq\eqlabel{PvDef}
P(v) =\sqrt{ \frac{\det T}{(2\pi)^{N_v}} } \, e^{- \frac{1}{2}v^t T v-B_v^t v - \frac{1}{2} B^t_v T^{-1} B_v}   \, \frac{\tilde\theta\left(B_h^t+v^t W  |Q\right)}{\tilde\theta\left(B^t_h -B_v^t T^{-1} W|Q-W^t T^{-1}W\right)}\,,
\eq
where $T$ and $Q$ are the connection matrices of the visible and hidden sectors, $W$ represent the inter-connections, $B_{v}$ and $B_{h}$ are the biases of the respective sector nodes, and $N_v$ is the number of visible nodes. The function $\tilde\theta$ is the Riemann-Theta function \cite{M1983} (with some implicit rescaling of the arguments), arising from the quantization of the hidden sector. It possesses intriguing mathematical properties, and appears in a diverse range of applications, including number theory, integrable systems, and string theory. As we will show in this work, this parametric density can in fact be used to model quite general densities of a given dataset via a maximum likelihood estimate of the parameters.

In our case the hidden sector of the Boltzmann machine is not binary, hence the conditional probabilities $P(h_i|v)$ are not well suited to be taken as feature vectors in a setup similar to \cite{HS2006}. We propose to use instead the conditional expectation of the hidden units, referred to as $E(h_i|v)$. The expectation can again be calculated explicitly, reading
$$
E(h_i|v)=-\frac{1}{2\pi\ii}\frac{\nabla_i\tilde\theta(v^t W+B^t_h|Q)}{\tilde\theta(v^t W+B^t_h|Q)}\,,
$$
where $\nabla_{i}$ denotes the $i$th inner derivative of the Riemann-Theta function.\footnote{Note that the imaginary prefactor arises from our chosen conventions. The final expressions for $E(h_i|v)$ are purely real for the parameter domains under consideration.} If we take $v\in\mathbb R^{N_v}$ and have $N_{h}$ hidden units, then 
\beq
\eqlabel{Emap}
E:\, \mathbb R^{N_v}\overset{W}{\longrightarrow} \mathbb R^{N_h} \overset{\nabla\tilde\theta/\tilde\theta}{\longrightarrow} \mathbb R\,.
\eq

We can view $E(h_i|v)$ as the $i$th activation function of a layer of $N_{h}$ units in a feedforward neural network. These layers can be arbitrarily stacked and combined with ordinary neural network layers. For these layers the network will learn not only the weights and biases of the linear input map, but also the parameter matrix $Q$ (for instance via gradient descent). That is, the form of the non-linearity most suitable for each node is learned from the data, in addition to the linear maps. Thus, such a unit is expected to possess a greater modeling capacity than a fixed standard neural network non-linear unit. An additional benefit of this novel unit is its robustness against normalization of inputs due to its intrinsic periodicity, \cf, figure \ref{EhvPlots}.  

Unfortunately, the explicit computation of the Riemann-Theta functions at each optimization step comes with a large overhead compared to usual non-linearities, \cf, section \ref{RTBMtheory}. Therefore an efficient implementation of the Riemann-Theta function is desirable. This work uses the explicit implementation in \cite{CK2017}, with \cite{R2017} as the math backend (which is based on an optimized implementation of \cite{DHBHS2003,SD2016}). At least for toy examples we find indications that smaller network sizes than for standard neural networks are sufficient, thereby raising the hope that the computational overhead stays managable.

This work is mainly about the theoretical foundations of the Riemann-Theta Boltzmann machine and the derived feedforward neural network. Though we give a couple of illustrative and explicit examples, we postpone a detailed study of applications to another time. It is astonishing that the mathematically complicated density \req{PvDef} can be trained successfully. 

For practical applications it would be desirable to better understand good parameter initializations. In this work we simply initialize via uniformly sampling the parameters from some fixed range. Introducing regularization, for example via Dropout \cite{SHKSS2014} would also be useful. On the implementation side it would be very desirable to implement the Riemann-Theta function more efficiently, perhaps using a GPU \cite{W2013}.

The outline is as follows. In section \ref{RTBMtheory} we will derive the Riemann-Theta Boltzmann machine in detail, laying the foundation for the following sections. The RTBM can be explicitly used to learn probability densities, as we will show in section \ref{RTBMmodelSec}. We introduce feedforward networks of expectation units in section \ref{TNNsection} and apply them to some simple toy examples. In section \ref{RTBMclassSec} we show how RTBMs can be used as feature detectors. The appendix collects some additional material: A derivation of the probability density of the continuous Boltzmann machine in appendix \ref{CBM}, the gradients needed for gradient descent in appendix \ref{gradientSection}, and the first two moments of $P(v)$ in appendix \ref{MomentsSec}.

\section{RTBM theory}
\label{RTBMtheory}

\begin{figure}
\begin{center}
  \includegraphics[scale=0.25]{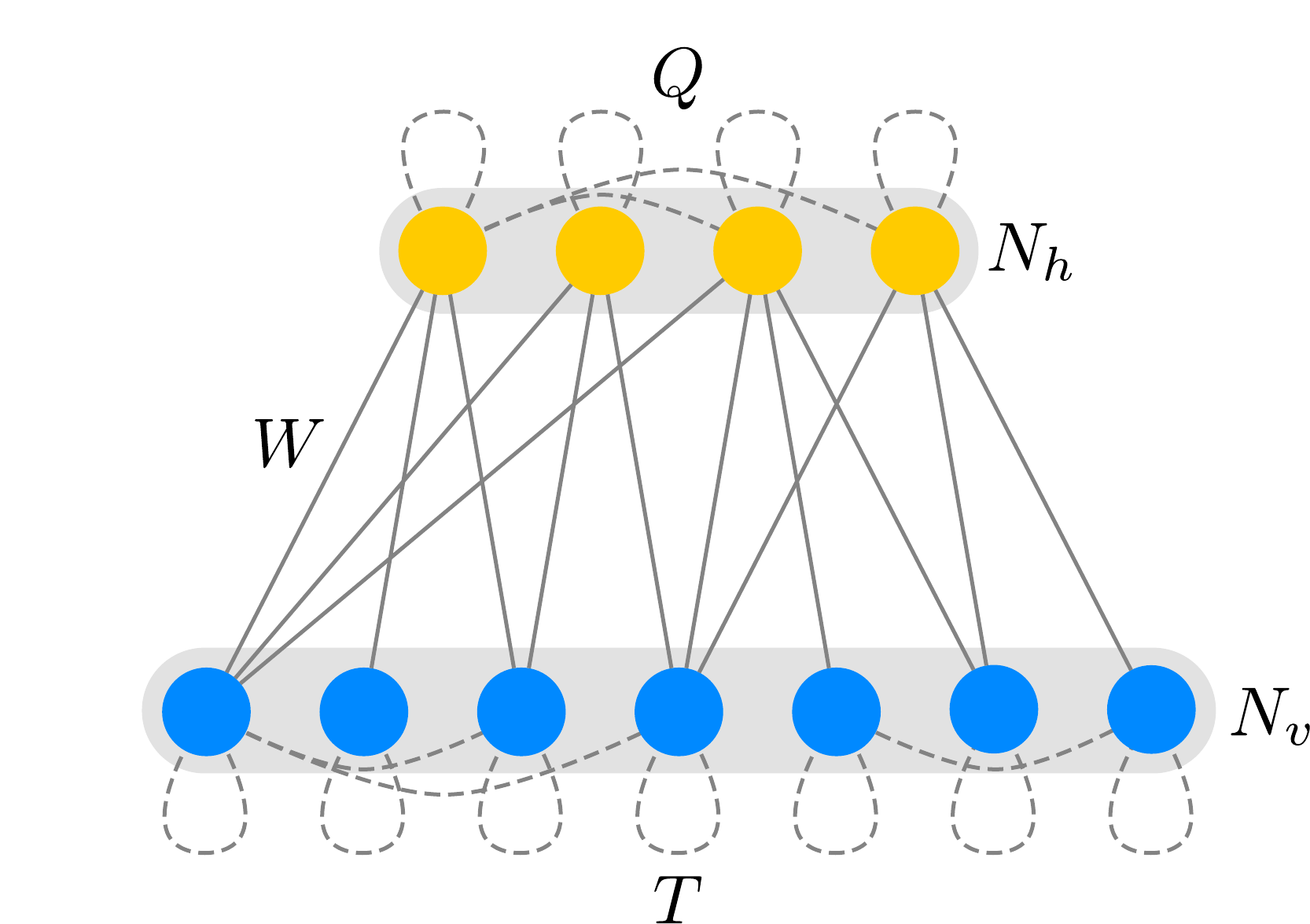}
  \caption{Illustration of the RTBM consisting of $N_{v}$ visible and $N_h$ hidden nodes. The two sectors are arbitrarly inter-connected with connection weights encoded in a matrix $W$ which can be either purely real or imaginary. The weights of the self-connections in a sector are encoded in matrices $Q$ and $T$ for the hidden sector and visible sector, respectively. These matrices have to be real, symmetric and positive definite.}
  \label{RTBMfig}
\end{center}
\end{figure}

\paragraph{The model}
We define a Riemann-Theta Boltzmann machine consisting of $N_v$ visible nodes and $N_h$ hidden nodes as follows. All nodes can be fully interconnected. The connection weights between the visible units are encoded in a real $N_{v}\times N_{v}$ matrix $T$, the weights of the interconnectivity of the hidden units in a real $N_{h}\times N_{h}$ matrix $Q$ and the connection weights between the two sectors in a $N_{v}\times N_{h}$ matrix $W$, which can be either purely real or imaginary. The setup is illustrated in figure \ref{RTBMfig}.
We combine the individual connectivity matrices into an overall connection matrix $A$ by defining $A$ as the block matrix
$$
A = \left(
\begin{matrix}
Q & W^t\\
W & T
\end{matrix}
\right)\,,
$$
of dimension $(N_v+N_h)\times (N_v+N_h)$. Let us restrict ourselves for the moment to the case with $W$ real. For reasons which will become more clear below, we require in this case that $A$ is positive definite.\footnote{Note that positive definiteness of $A$ requires that the diagonal elements of $T$ and $Q$ are greater than zero. Therefore self-couplings of the RTBM need to be present.}  The Schur complement $A/T$ of the block $T$ of $A$ is given by
\beq\eqlabel{SchurCompl}
A/T = Q - W^t T^{-1} W\,.
\eq
As $A$ is positive definite, so are $A/T$, $T$ and $Q$.

The states of the visible nodes are taken to be continuous in $\mathbb R$, while we restrict the states of the hidden nodes to be in $\mathbb Z$. Hence, we are constructing here a generalization of the Gaussian-Bernoulli Boltzmann machine (whose hidden states are only binary, \cf, \cite{HS2006,B2012,H2012}). We combine the two state vectors to a single vector $x$ as
$$
x = \left(
\begin{matrix}
h\\
v
\end{matrix}
\right)\,,
$$
and define the energy of the system to be 
$$
E(v,h) = \frac{1}{2} x^t A x+B^t x\,.
$$
The quadratic form reads
$$
x^t A x = v^t T v+h^t Q h + 2 v^t W h\,,
$$
and we introduced above an additional bias vector 
$$
B = \left(
\begin{matrix}
B_h\\
B_v
\end{matrix}
\right)\,.
$$
Note that the positive definiteness of $A$ ensures that $E > 0$ for large $x$.

The canonical partition function $Z$ of the system is obtained via integrating/summing over all states, \ie, 
\beq\eqlabel{Zdef}
Z=\int_{-\infty}^\infty [dv]\sum_{[h]} e^{-E(v,h)}\,\,,
\eq
where $[dv]$ stands for the measure $dv_1dv_2\dots dv_{N_v}$ and $[h]$ is an abbreviation of $h_1,h_2,\dots h_{N_h}$. 

\paragraph{Riemann-Theta function}
The key observation we make is that the summation over the hidden states in the partition function \req{Zdef} can be performed explicitly, in contrast to the case of an ordinary Boltzmann machine. In detail, as the energy $E(v,h)$ is a quadratic form, the summation over the discrete states corresponds to a Riemann-Theta function, defined as \cite{M1983}
\beq\eqlabel{RTdef}
\theta(z|\Omega) = \sum_{n\in \Z^g} e^{2\pi\ii\left(\frac{1}{2} n^t\Omega n+n^t z\right)}\,,
\eq
with $\Omega$ a matrix whose imaginary part is positive definite. The above series converges absolutely and uniformly on compact sets of the $z$ and $\Omega$ spaces, as long as the imaginary part of $\Omega$ is positive definite. The Riemann-Theta function is symmetric in $z$ and quasi-periodic, \ie,
\beq
\theta(z+m_1+\Omega m_2|\Omega) = e^{-2\pi\ii \left(\frac{1}{2} m_2^t\Omega m_2+m_2 z \right)} \theta(z|\Omega)\,,
\eq
with $m_1,m_2\in \Z^g$. Note that the set of points $m_1+\Omega m_2$ forms a $g$-dimensional period lattice. The Riemann-Theta function also possesses a modular transformation property, however, this will not be of relevance for this work. 

Riemann-Theta functions commonly arise in integrable differential equations with applications in diverse areas. For example, the Kadomtsev-Petviashvili equation, which describes the propagation of two-dimensional waves in shallow water, possesses a class of quasi-periodic solutions in terms of the Riemann-Theta function. 

Though a simple observation, the linkage between the stochastic network Boltzmann machine variant introduced above and Riemann-Theta functions, is novel and intriguing. 

\paragraph{Partition function}

Let us define a free energy $F$ as
\beq\eqlabel{Fdef}
F(v) = -\log \sum_{[h]} e^{-E(v,h)}\,,
\eq
such that
$$
Z = \int [dv]\, e^{-F(v)}\,.
$$
We can immediately write down a closed form expression for the free energy in terms of the Riemann-Theta function introduced above, as the summation over the states $h$ corresponds to a summation over an $N_{h}$-dimensional unit lattice and the energy $E$ is a quadratic form:
\beq\eqlabel{Fv}
F(v) = \frac{1}{2}v^t Tv + B^t_v v -\log \tilde\theta\left(v^tW+B_h^t|Q\right)\,,
\eq
where we made use of the symmetry $\theta(z|\Omega) = \theta(-z|\Omega)$ and defined
$$
\tilde\theta(z|\Omega):=\theta\left(\frac{z^t}{2\pi\ii}\right|\left.\frac{\ii\Omega}{2\pi}\right)\,.
$$
Note that the redefined $\theta$ has periodicity $\tilde\theta(z+2\pi\ii\, n|\Omega)=\tilde\theta(z|\Omega)$, with $n$ a vector of integers and that we will also sometimes refer to $g=N_{h}$ as genus. For $g=1$ the function \req{RTdef} is also known as the 3$rd$ Jacobi-Theta function.

The partition function $Z$ can be calculated explicitly in a similar fashion. First we integrate out the visible sector, making use of the gaussian integral 
\beq\eqlabel{gaussint}
\int [dx]\,e^{-\frac{1}{2}x^t Q x + y^t x } = \frac{(2\pi)^{N/2}}{\sqrt{\det{Q}}}\, e^{\frac{1}{2} y^t Q^{-1} y}\,,
\eq
this yields
$$
I(h) = \frac{(2\pi)^{N_v/2}}{\sqrt{\det T}}\, e^{-\frac{1}{2}h^tQh-B^t_h h +\frac{1}{2}(h^tW^t+B^t_v)T^{-1}(W h + B_v)}\,.
$$
Subsequently, we perform the summations over $h$, yielding the final expression
$$
Z = \frac{(2\pi)^{N_v/2}}{\sqrt{\det T}}\,e^{\frac{1}{2} B^t_v T^{-1} B_v}\, \tilde\theta\left(B^t_h -B_v^t T^{-1} W|Q-W^t T^{-1}W\right)\,.
$$

\paragraph{Probability density}

The probability that the system will be in a specific state is given by the Boltzmann distribution 
\beq\eqlabel{BoltzmannP}
P(v,h) = \frac{e^{-E(v,h)}}{Z}\,.
\eq
Marginalization of $h$ yields the distribution for the visible units, \ie,
$$
P(v) = \frac{e^{-F(v)}}{Z}\,,
$$
with the free energy as defined in \req{Fdef}. As we have closed form expressions for both $Z$ and $F$, we can immediately write down the closed form solution
\beq\eqlabel{Pv}
P(v) =\sqrt{ \frac{\det T}{(2\pi)^{N_v}} } \, e^{- \frac{1}{2}v^t T v-B_v^t v - \frac{1}{2} B^t_v T^{-1} B_v}   \, \frac{\tilde\theta\left(B_h^t+v^t W  |Q\right)}{\tilde\theta\left(B^t_h -B_v^t T^{-1} W|Q-W^t T^{-1}W\right)}\,.
\eq
We observe that $P(v)$ consists of a multi-variate Gaussian for the visible units with a visible unit dependent prefactor given by a Riemann-Theta function. This probability distribution for the visible units of the RTBM is one of the core results of this work. Note that this density is well-defined for $T$, $Q$ and $A/T$ positive definite (\cf, \req{SchurCompl}), which explains why we required above $A$ to be positive definite. Furthermore, in order that $P(v)$ is real, we take these matrices to be real, and $B_v$ as well. The coupling matrix $W$ and the bias $B_h$ can then be chosen either both from the real (phase I) or the imaginary (phase II) axis, giving rise to a two phase structure connected at the null-space of $W$ and $B_{h}$. The realness of $P(v)$ in phase II follows from the fact that in the Riemann-Theta function summation \req{RTdef} the imaginary parts cancel out between terms with $n$ reflected at the origin.

\begin{figure}
\begin{center}
  \includegraphics[scale=0.33]{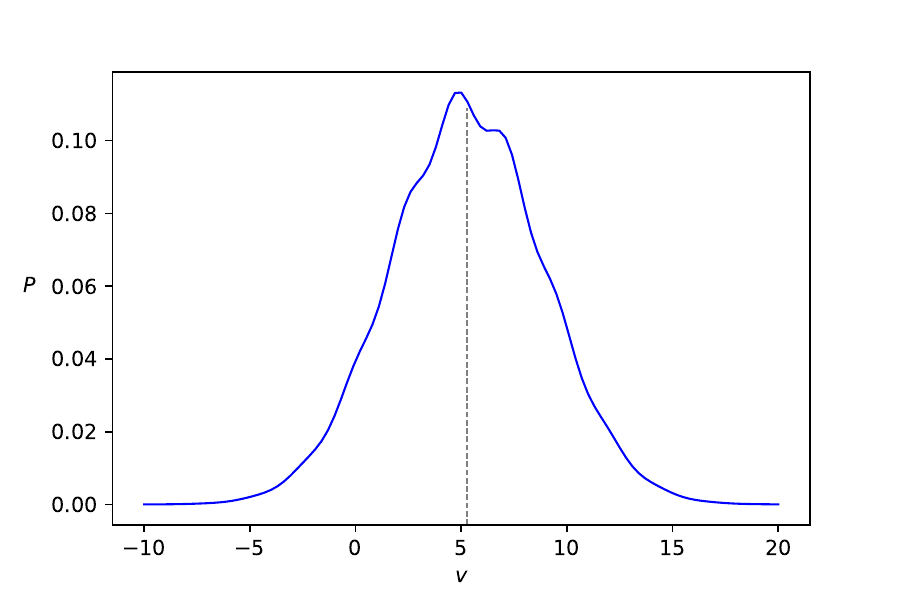}
  \includegraphics[scale=0.33]{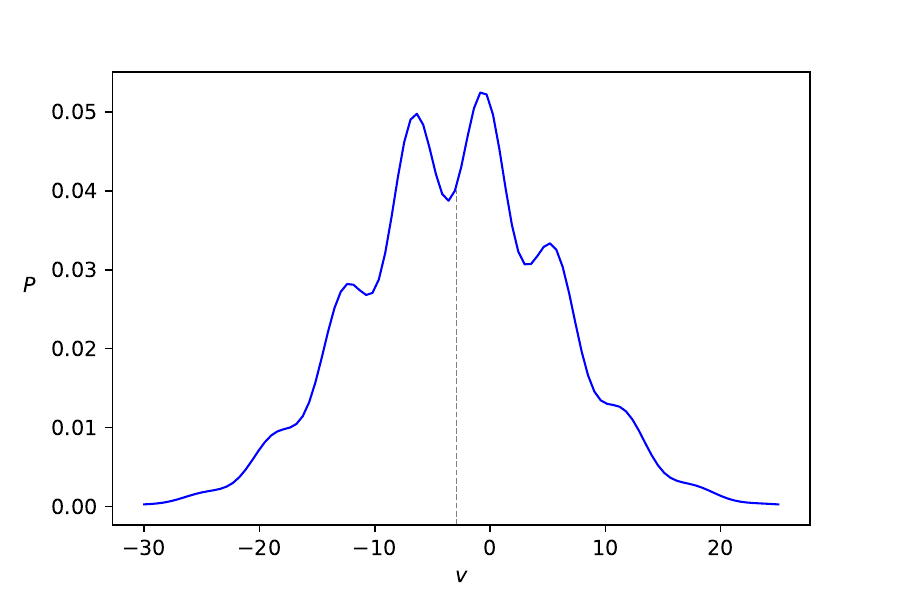}
  \includegraphics[scale=0.33]{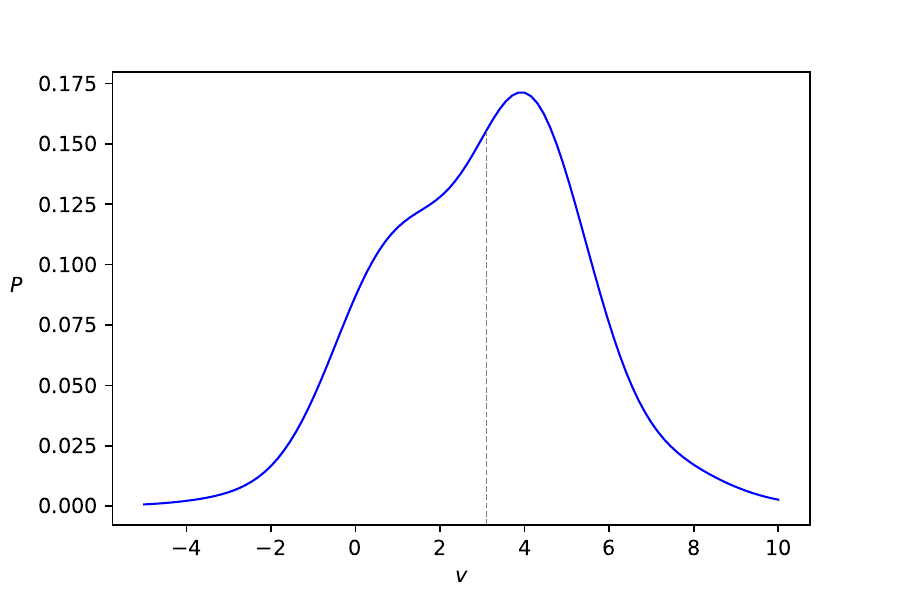}
  \includegraphics[scale=0.33]{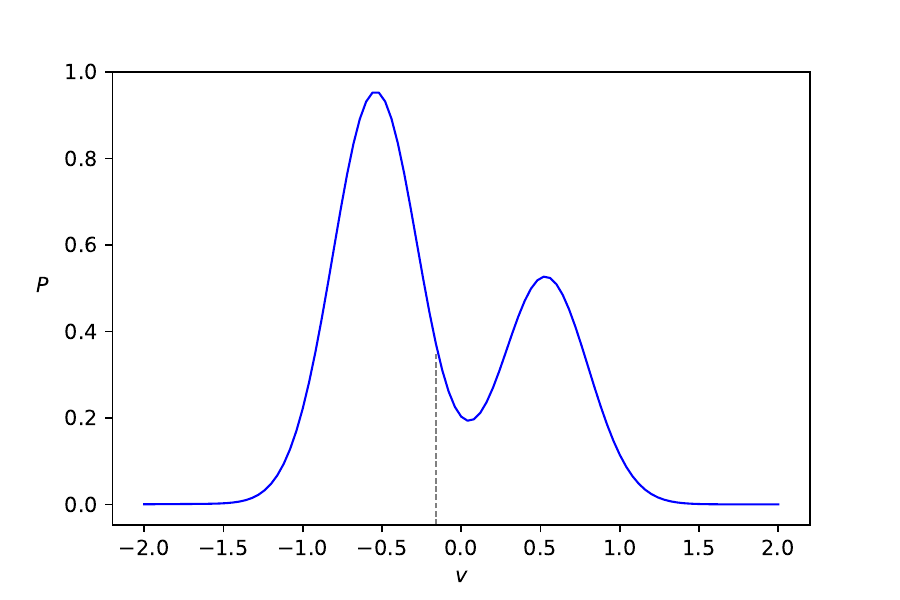}
  \includegraphics[scale=0.33]{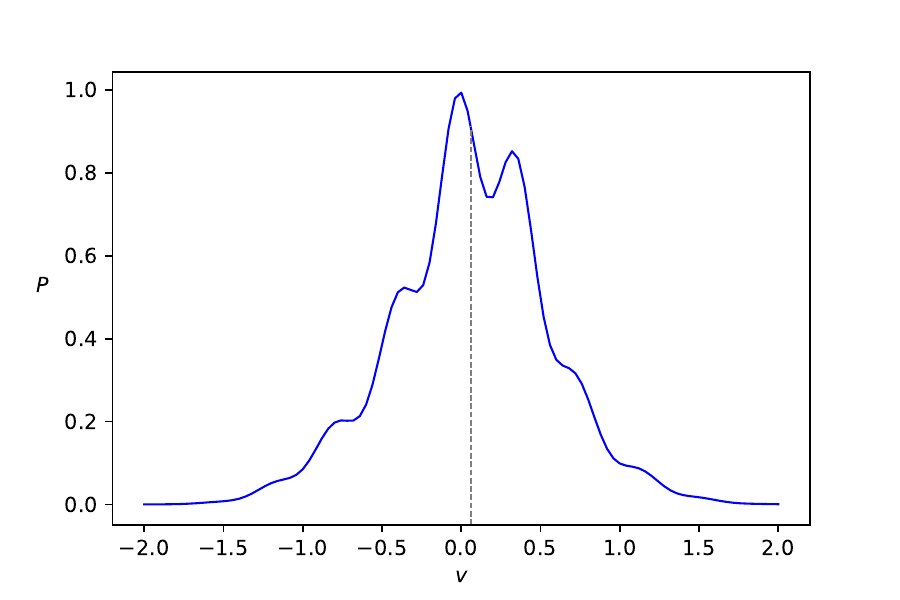}
  \includegraphics[scale=0.33]{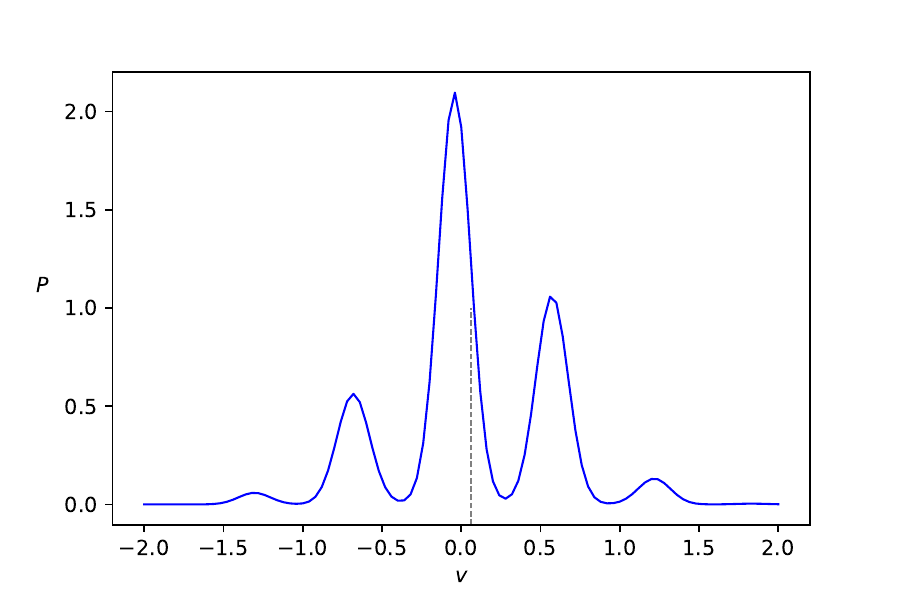} 
  \caption{Plots of $P(v)$ for one-dimensional $v$ and various choices of parameters $B_v,B_h,W,T,Q$. Top row: Phase I. Bottom row: Phase II. The first examples in both phases are with one-dimensional Q, while the remaining plots are for $Q$ of size $2\times 2$. The gray dashed lines mark the means of the distributions.}
  \label{PvPlots}
\end{center}
\end{figure}

For illustration, some plots of $P(v)$ in the $N_v=1$ case for a sample choice of parameters are given in figure \ref{PvPlots}. 
We observe that for appropriate choices of parameters non-trivial generalizations of the Gaussian are obtained. Note that the moments of the probability density \req{Pv} can be easily calculated: see appendix \ref{MomentsSec}.

It is illustrative to consider the logarithmic probability $\log P(v)$ in the case with diagonal $Q$. For such $Q$ we have the factorization
\beq\eqlabel{ThetaDiagFactorization}
\theta(z|\Omega) = \prod_{i=1}^{N_h} \theta(z_i|\Omega_{ii})
\eq
(the Riemann-Theta function factorizes into Jacobi-Theta functions). Hence,
$$
\log P(v) = -\frac{1}{2}v^t T v -B^t_v v +\sum_{i=1}^{N_h} \log \tilde\theta\left(\left(B_h^t+v^t W\right)_i  |Q_{ii}\right) +{\rm const}\,.
$$
The constant term includes the Riemann-Theta function of the normalization, which is however not factorizable for generic parameters. In particular, this means that a restricted version of the Riemann-Theta Boltzmann machine with $Q$ diagonal does not provide striking advantages from a computational point of view.

The zeros of the redefined Jacobi-Theta function $\tilde\theta(z)$ (given in \req{RTdef} with $g=1$) are located at $z_*=-\left(n-\frac{1}{2}\right)\Omega +\pi\ii $. Hence, we infer from the definition \req{RTdef} that $\log P(v)$ is well defined, as $\tilde\theta(z|\Omega)>0$ always holds in the parameter spaces under consideration. In phase II the $\log\tilde\theta$ terms are periodic in $v$. Hence in this case the logarithmic density consists of an overlap of an inverse paraboloid and $N_{h}$ periodic functions. The interpretation of phase I is less clear as $\tilde\theta$ is not periodic in $v$. Nevertheless, via proper tuning of parameters, suitable solutions can be found, \cf, figure \ref{PvPlots}. An interpretation can be given as follows. From the property of the Riemann-Theta function
\beq\eqlabel{ThetaFundRel2}
\theta(z+ \Omega n |\Omega)= e^{2\pi\ii\left(-n^t z -\frac{1}{2}n^t\Omega n\right)}\theta(z|\Omega)\,,
\eq
where $n\in \mathbb Z^g$, we deduce that
\beq\eqlabel{LogThetaPeriodRelation}
\log\tilde\theta(B_h^t+v^t W -\Omega\, n |\Omega)=  \log\tilde\theta(B_h^t+v^t W |\Omega)-n^t(B_h^t+v^t W)+\frac{1}{2}n^t \Omega n\,.
\eq
Hence, $P(v)$ can be seen as a quadratic surface overlapped with periodic functions.

A remark is in order here. The zero locus of the Riemann theta function is given by an analytic variety of complex dimension $g-1$. In cases where the symmetric matrix $\Omega$ is obtained from a genus $g$ \textit{Riemann surface} by period integrals of its holomorphic one-forms, the zero locus of the Riemann theta function is exactly determined by the so called \textit{Riemann vanishing theorem} (see \cite{FK1980} for further details). However, in general this is not always the case. The reason is that the dimension of the space of $\Omega$'s, known as the Siegel upper half space, is that of symmetric matrices and therefore grows like $g (g+1)/2$, whereas the dimension of the moduli space of genus $g$ Riemann surfaces is zero for $g=0$, one for $g=1$ and $3g-3$ for $g > 1$. As one can easily check, these two dimensions only match for $g < 4$, and for all other cases the number of parameters of the Siegel upper half space is bigger than that of the Riemann surface. Therefore, in general the zero set of the Riemann theta function is not known explicitly, and its study is an important topic in current mathematics. For the $P(v)$ studied in this paper, these considerations are not of utmost relevance, since from the definition of the partition function \req{Zdef} it is clear that for real parameters $Z$ only vanishes for $E(v,h)=\infty$ and therefore $P(v)$ is well defined in phase I, as long as the parameters are finite and satisfy the positive definiteness conditions above. For phase II the absence of zeros is less clear. However, after studying some concrete examples we observed that the gradient flow in parameter space usually does not seem to encounter such points.

\paragraph{Conditional density}

The conditional probability for the hidden units is given by 
$$
P(h|v) = \frac{P(v,h)}{P(v)} =  \frac{e^{-\frac{1}{2}h^t Q h-(v^t W+B_h^t) h}}{\tilde \theta(v^t W+B^t_h|Q)}\,.
$$
Note that $P(h|v)$ is independent of $T$ and $B_v$. For diagonal $Q$, the density can be factorized, \ie,
\beq\eqlabel{PhvFactorization}
P(h|v) = \prod_{i=1}^{N_h}P(h_i|v)\,.
\eq
In contrast to the ordinary Boltzmann machine, here we have infinitely many different states of the hidden units. Hence, it is useful to consider the expectation $E(X|Y):=\sum_X P(X|Y)X$ of the $i$th hidden unit state. Taking the expectation and marginalization of the remaining components of $h$ yields the expression
\beq\eqlabel{EhvDef}
E(h_i|v) = \frac{1}{\tilde \theta(v^t W+B^t_h|Q)}\sum_{[h]} h_i \, e^{-\frac{1}{2}h^t Q h-(v^t W+B^t_h) h}\,.
\eq
Comparing with the definition \req{RTdef} and equation \req{Fv}, we infer the relation
$$
E(h_i|v)= -\frac{\partial  F(v)}{\partial (B_h)_i}\,.
$$
Taking the derivative yields
\beq\eqlabel{Eunit}
E(h_i|v)=-\frac{1}{2\pi\ii}\frac{\nabla_i\tilde\theta(v^t W+B^t_h|Q)}{\tilde\theta(v^t W+B^t_h|Q)}\,,
\eq
where $\nabla_{i}\theta$ denotes the $i$th directional derivative of the first argument of the Riemann-Theta function. For diagonal $Q$, $E(h_i|v)$ reduces via the factorization property \req{ThetaDiagFactorization} to
\beq\eqlabel{EdDef}
E_d(h_i|v) = -\frac{1}{2\pi\ii}\frac{\tilde \theta'(\left(v^t W+B^t_h\right)_i|Q_{ii})}{\tilde \theta(\left(v^t W+B^t_h\right)_i|Q_{ii})}\,.
\eq
Here, $\theta'$ refers to the derivative with respect to the first argument. (We use a subscript $_{d}$ to indicate that this expression holds only in the diagonal case.) 

\begin{figure}
\begin{center}
  \includegraphics[scale=0.33]{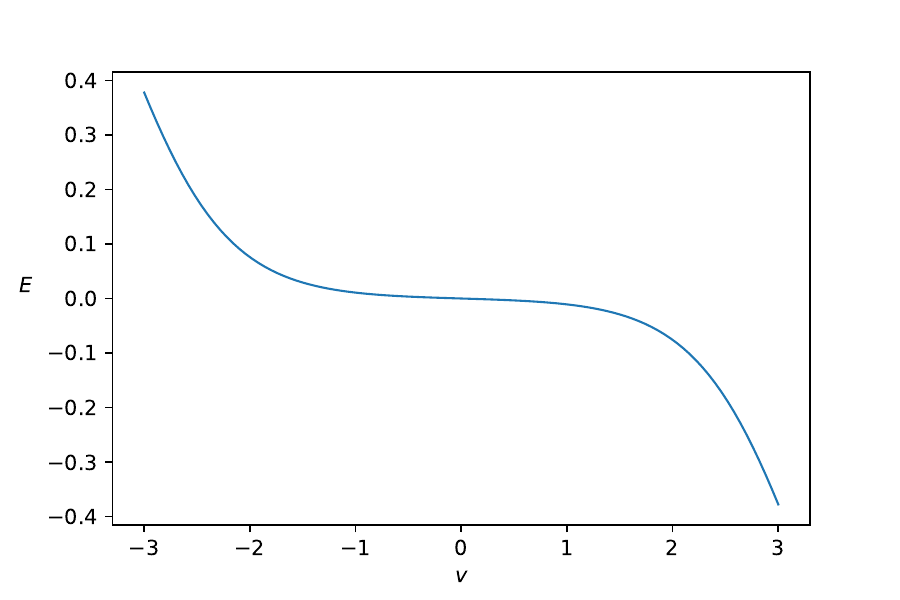}
  \includegraphics[scale=0.33]{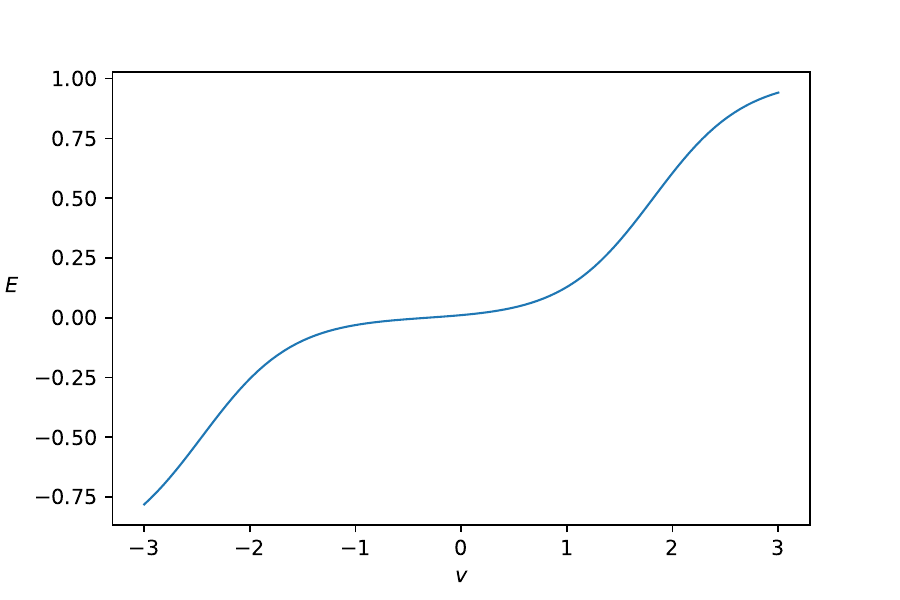}
  \includegraphics[scale=0.33]{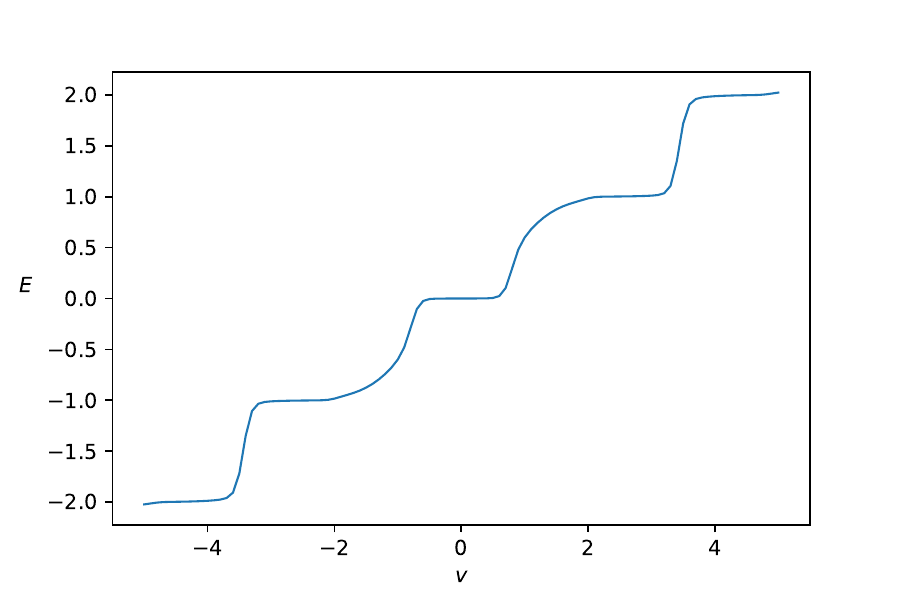}
  \includegraphics[scale=0.33]{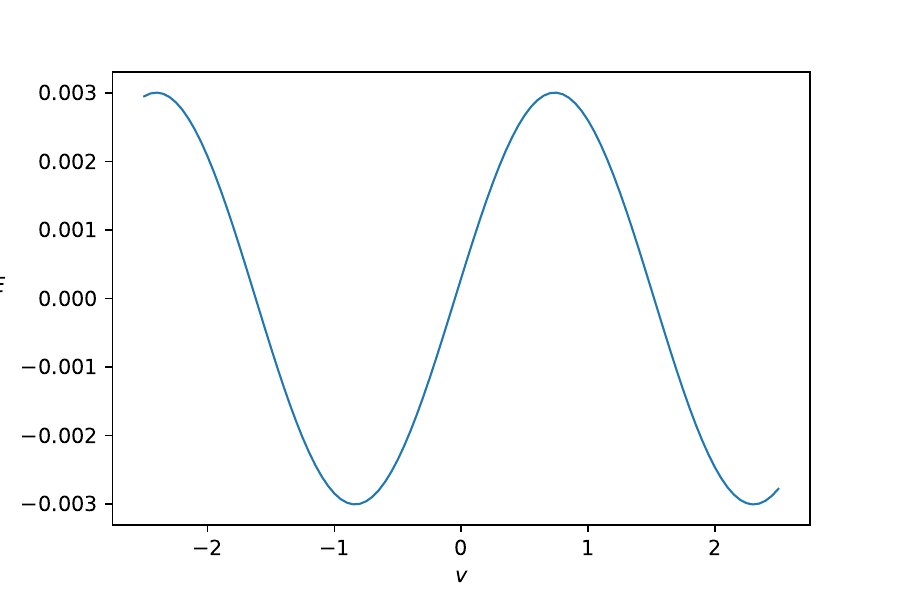}
  \includegraphics[scale=0.33]{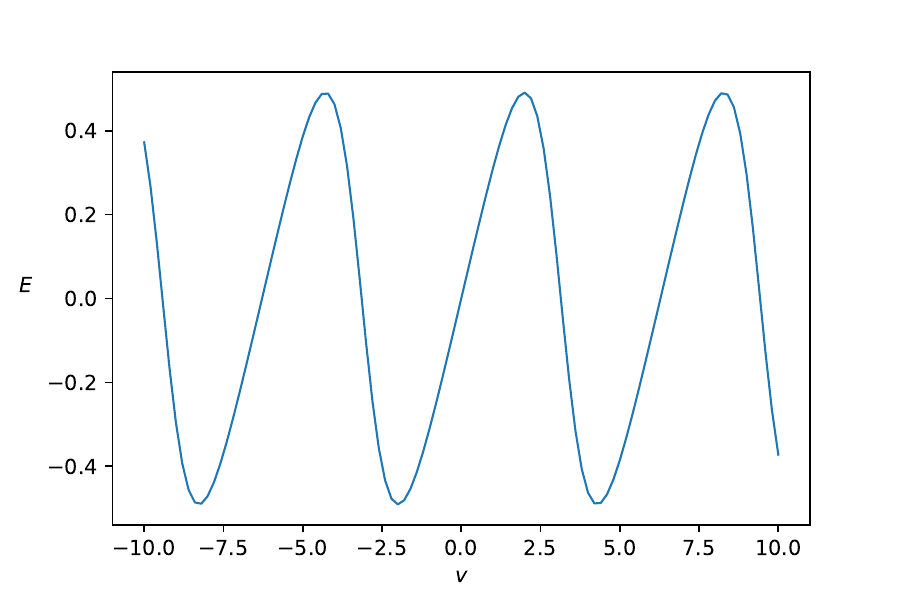}
  \includegraphics[scale=0.33]{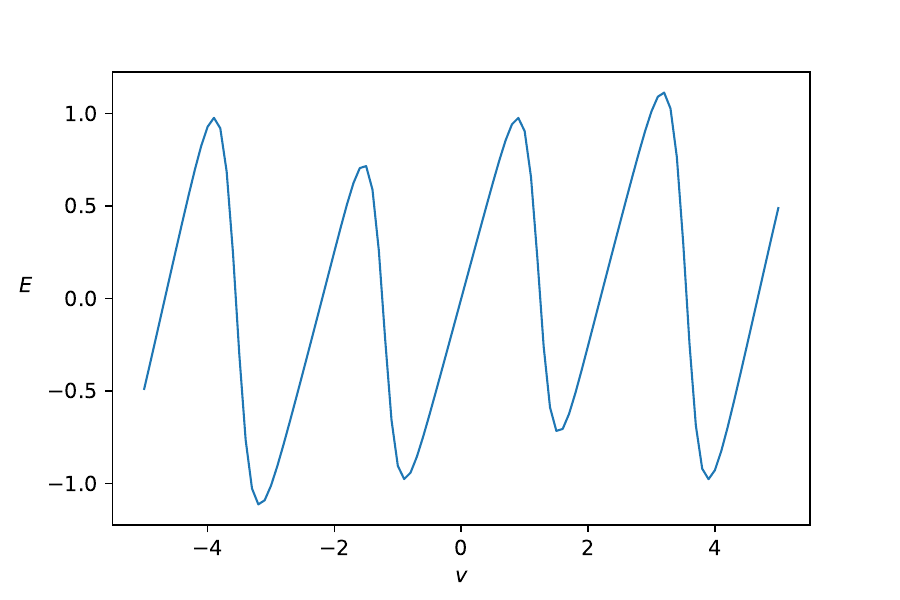} 
  \caption{Plots of $E(h|v)$ for one-dimensional $v$ and various choices of parameters $B_h,W,Q$. Top row: Phase I. Bottom row: Phase II. The plots on the RHS are with $Q$ of size $2\times 2$, whereas the remaining plots are with one-dimensional $Q$.}
  \label{EhvPlots}
\end{center}
\end{figure}

It is illustrative to consider the diagonal case in more detail. Clearly, in phase II the expectation $E_d$ is periodic in $v$ due to the known relation $\frac{\theta'(z+\pi|\Omega)}{\theta(z+\pi|\Omega)}=\frac{\theta'(z|\Omega)}{\theta(z|\Omega)}$ \cite{WW1990}. In contrast, in phase I it is not periodic, but is rather some trending periodic function. This can be inferred from \req{LogThetaPeriodRelation}, which turns under the derivative into
\beq\eqlabel{EhvTrend}
\partial_{(B_{h})_i}\log\tilde\theta(B_h^t+v^t W -\Omega\, n |\Omega)=\partial_{(B_{h})_i}\log\tilde\theta(B_h^t+v^t W |\Omega) - n_i\,.
\eq
The different behaviors of $E_{d}$ in the two phases is illustrated using a sample choice of parameters in figure \ref{EhvPlots}.\footnote{Note that in phase II the expectations are purely imaginary and that we take the freedom to rotate to the real axis in this case.}

\paragraph{Learning}

The learning of parameters of \req{PvDef} is performed via maximum likelihood. That is, for $N$ samples $x_{i}$ of some unknown probability density we take the cost function
$$
\mathcal C = -\sum_{i=1}^N \log P(x_i)\,,
$$
and solve the optimization problem
$$
\underset{B_v,B_h,W,T,Q}{\rm argmin}\, \mathcal C\,.
$$
The gradients of $P$ are easy to calculate, \cf, \ref{PvGradients}. Hence, we can solve the optimization problem either via a gradient or non-gradient based technique.  In this work, we will mainly make use of the CMA-ES optimizer \cite{hansen2001ecj}, which follows an evolutionary strategy and is suitable for $N_{h}$ small ($N_{h} < 5$).

The parameters of $P(v)$ need to satisfy the condition that $A/T$, $T$ and $Q$ are positive definite. Finding an initial solution to these conditions can be easily achieved by generating a random real matrix $X$ of size $(N_v+N_h)\times(N_v+N_h)$ and taking $A=XX^{t}$. For all examples presented in this paper the $X$ matrix elements are sampled from a uniform distribution in the $[-1,1]$ domain. The component matrices can then be directly extracted from $A$ and will automatically fulfill the above conditions. However, what is less clear is that during the optimization, we stay in the allowed parameter regime.\footnote{Note that in this work we optimize directly the elements of $A$ and therefore we may loose the positive-definiteness. An alternative approach would be to optimize over the elements of $L$ given by the log-Cholesky decomposition of $A=LL^T$, which would guarantee positive-definiteness.} We observe empirically that this is often the case, \ie, the parameter flow seems to tend to conserve the conditions, at least for $N_{h}$ small. Note that for CMA-ES a tuning of the initial standard deviation to be sufficiently small may be required. In case we encounter a bad solution candidate with CMA-ES, \ie, not satisfying the positive definiteness condition on $A$, the method is set up to replace the bad solution with a new solution candidate until the total desired population size for each iteration step is reached. For increasing $N_{h}$ we expect to hit more frequently inconsistent solutions, therefore it would be promising to switch to an optimization algorithm taylored for positive definiteness constraints.

Finally, we remark that a suitable initial $Q$ value is needed for convergence to a good solution. At the time being we only have indirect control over the $Q$ initialization via the range of allowed values for the $X$ entries and the CMA-ES range bound.

\paragraph{Computation}
The main computational cost of the Riemann-Theta Boltzmann machine lies in the evaluation of the Riemann-Theta function and its derivatives, as the complexity scales exponentially with the number of hidden units. One should take note that the current implementation of the Riemann-Theta function allows us to experiment on a desktop computer only with 3 to 4 hidden units comfortably.

The algorithm to evaluate the Riemann-Theta function can be sketched as follows. The evaluation requires the identification of the shortest vector in a given $N_{h}$-dimensional lattice, which is known as the shortest vector problem (SVP), \cf, \cite{MIC2001} and references therein. The Lenstra-Lenstra-Lovasz algorithm (LLL) can be used to solve approximately for the shortest vector in polynomial time with the error growing exponentially with increasing $N_{h}$. As in practice the error grows only slowly for $N_{h}$ not too large, the LLL algorithm is sufficient for our purposes. See \cite{FJK2017} for an extended discussion about computational aspects of the Riemann-Theta function. 

In this work we make use of the algorithm of \cite{DHBHS2003} for the computation of the Riemann-Theta function and its derivatives. This algorithm is fully vectorizable over the set of input data, which means that for given parameters $T,W,Q$ we can calculate the Riemann-Theta function and its derivatives for the complete input dataset at once, with no significant added cost in comparision to a single evaluation.

\section{RTBM mixture models}
\label{RTBMmodelSec}

\begin{figure}
\begin{center}
  \includegraphics[scale=0.25]{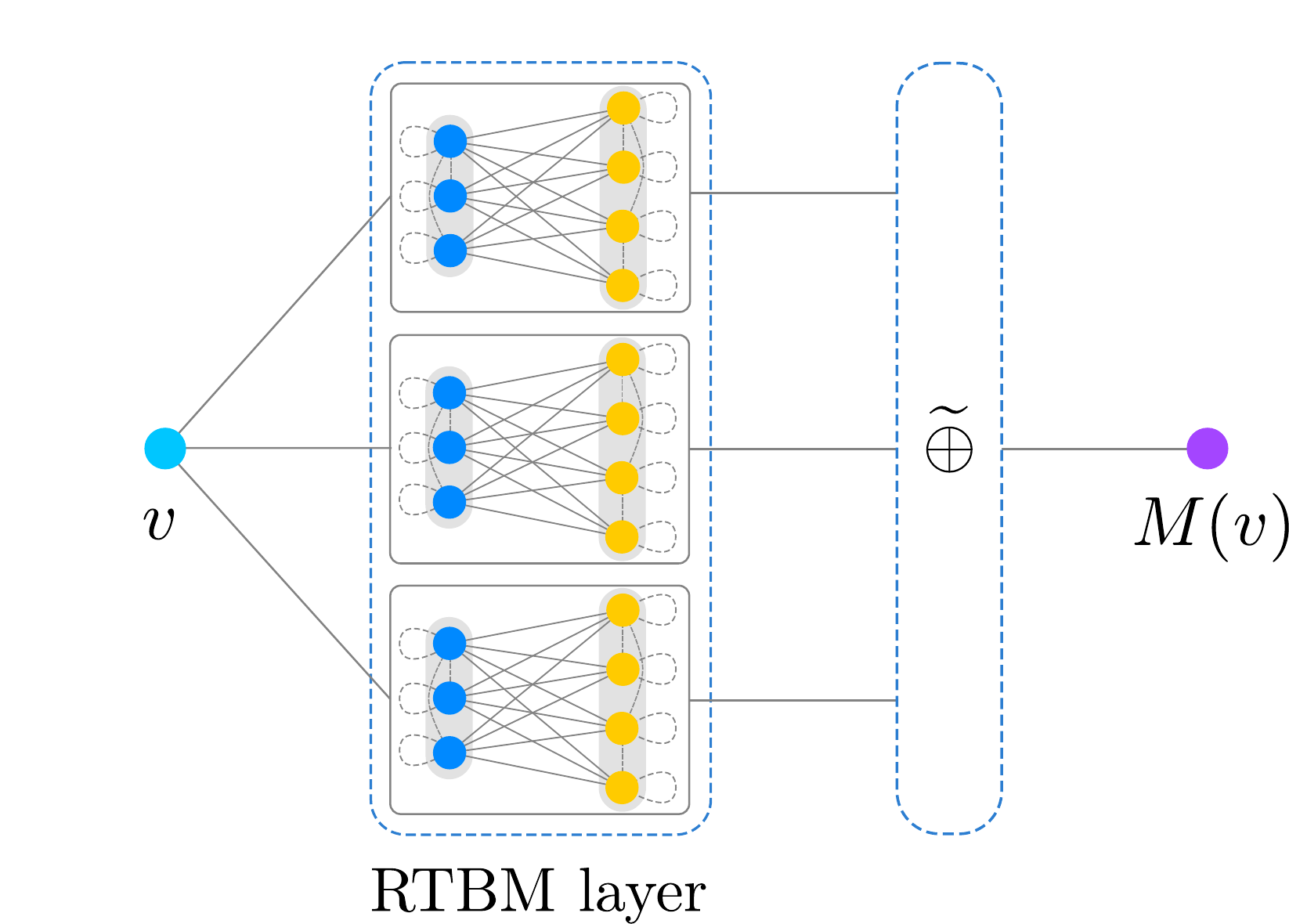}
  \caption{Illustration of the RTBM mixture model. The inputs are feed into a layer of RTBMs. The outputs of the RTBMs are then fed into a layer which adds and normalizes according to \req{MvDef}.}
  \label{RTBMmixturefig}
\end{center}
\end{figure}

\paragraph{Density estimation}
We saw in the previous section that the probability density of the RTBM visible units is a non-trivial generalization of the Gaussian density. Hence, we expect that for $N_{h}\rightarrow\infty$, and considering a mixture model 
\beq\eqlabel{MvDef}
M(v) = \frac{1}{\sum_{i=1}^N e^{\omega_i}} \sum_{i=1}^N e^{\omega_i}\, P^{(i)}(v)\,,
\eq
with $N$ the number of components, we can approximate a given smooth probability density arbitrarily well, as long as the density vanishes at the domain boundaries. Note that the $P^{(i)}$ should be centered at the degenerate or far separated maxima and that the exponential weighting in \req{MvDef} ensures that $M(v) \geq 0$ for all $\omega_{i}$. The mixture model setup is illustrated in figure \ref{RTBMmixturefig}.

It is well known that ordinary mixture models with components based on standard distributions, like the Gaussian, are well suited to model various kinds of low dimensional probability densities for sufficiently large $N$. However, for generic target distributions and finite $N$ good results are not always to be expected. Using neural networks instead to model probability density functions comes with the advantage of a high modeling capacity, but with the drawback that it is difficult to obtain a normalized output, \cf, \cite{modha1994,L2001,GJ1998}. The benefit of taking the RTBM density function as components of a mixture model, as in \req{MvDef}, is that we have the best of both worlds: an intrinsically normalized result and a high intrinsic modelling capacity. The learning of $M(v)$ is performed as for $P(v)$ described in the previous section.

\paragraph{Examples}
\label{Examplesec}
\begin{figure}[t!]
  \begin{center}
  \includegraphics[scale=0.33]{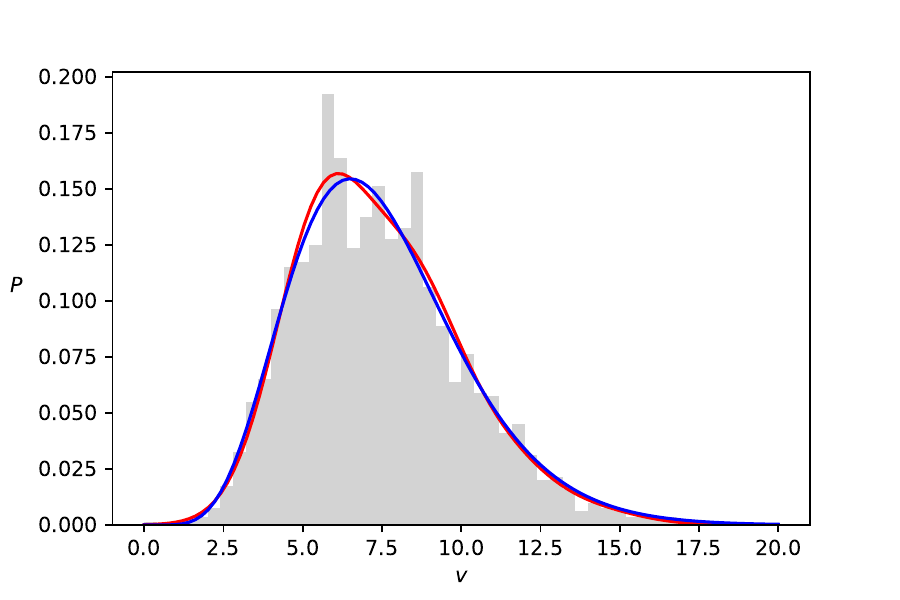}
  \includegraphics[scale=0.33]{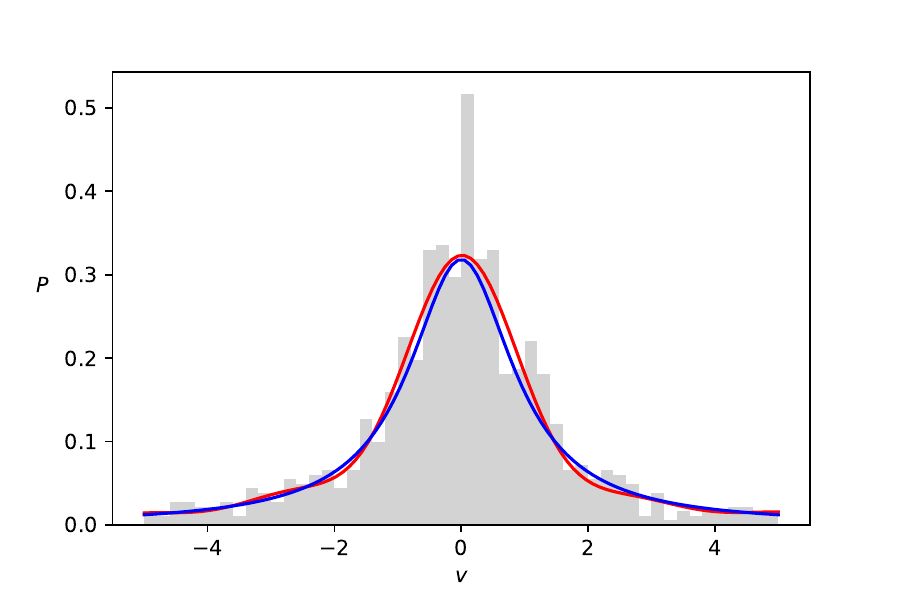}
  \includegraphics[scale=0.33]{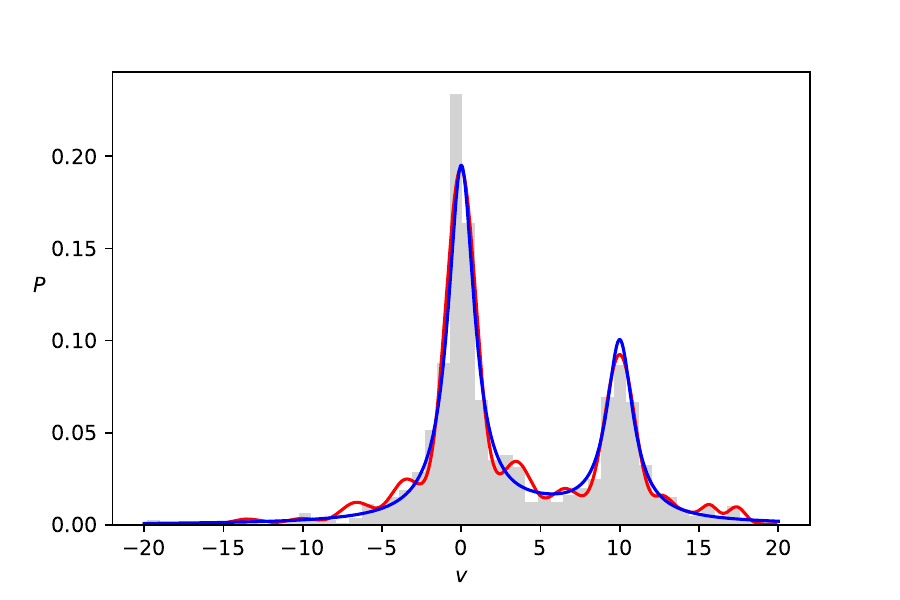}
  \includegraphics[scale=0.33]{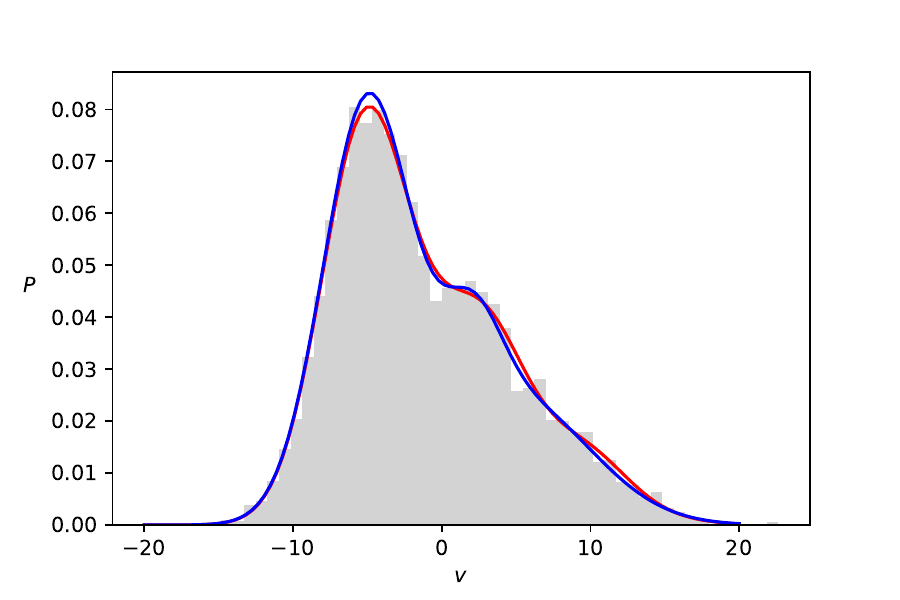}    
  \includegraphics[scale=0.33]{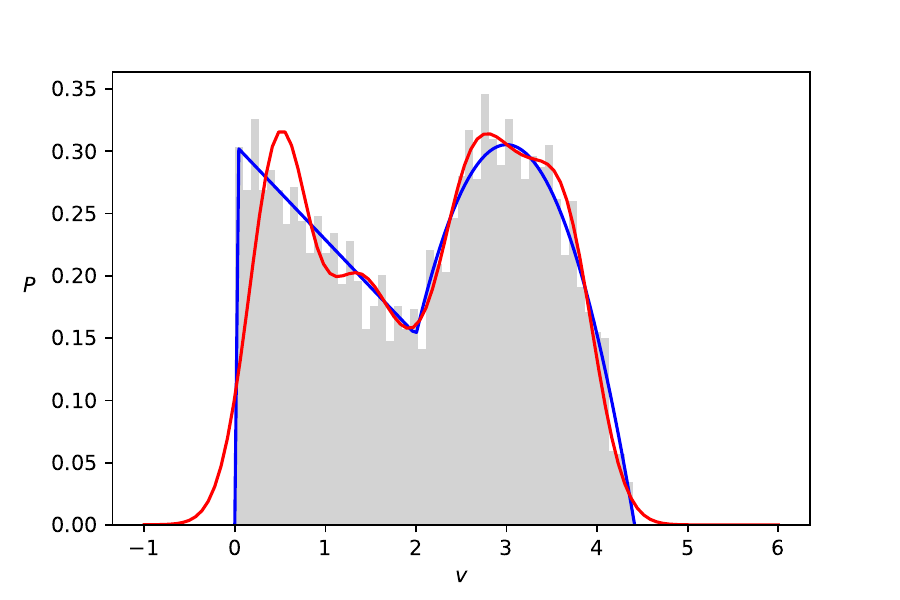}  
  \includegraphics[scale=0.33]{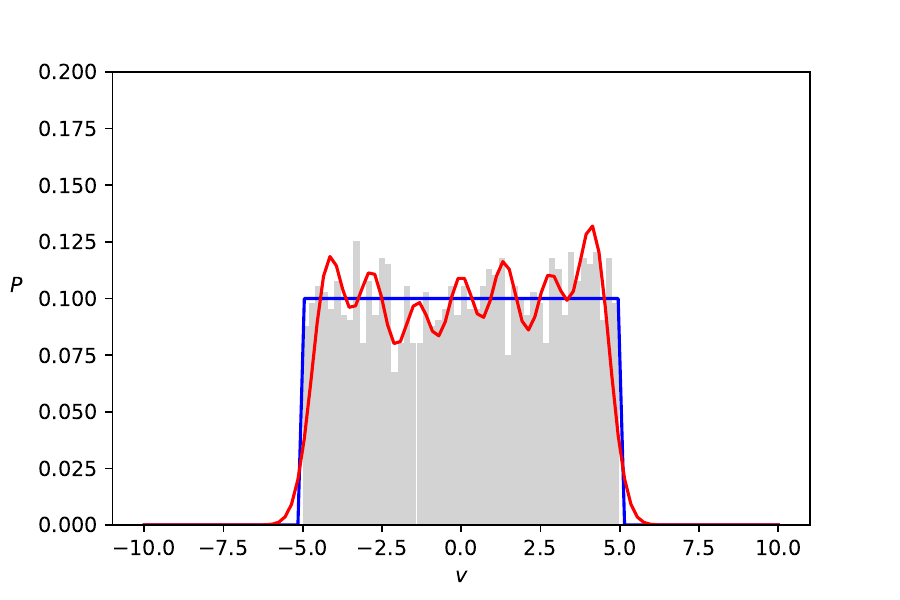}
  \caption{Top left: Gamma distribution fitted by a single RTBM with $N_{h}=2$.  Top middle: Cauchy distribution fit via a single RTBM with $N_{h}=3$. Top right: Fit of Cauchy distribution mixture via a layer of two RTBMs with $N_{h}=3$. Bottom left: Gaussian mixture fit by a single RTBM with $N_{h}=3$. Bottom middle: Custom mixture model fit using a single RTBM with $N_{h}=4$. Bottom right: Uniform distribution fit via a single RTBM with $N_h=3$. In all figures the blue line corresponds to the underlying true distribution, while the red line is the fit. The histograms show the samples the models are trained on.}
  \label{CauchyDistFig}
  \end{center}
\end{figure}
As a first example, let us consider the gamma distribution with probability density function reading
$$
p_\gamma(x,\alpha,\beta) = \frac{\beta^\alpha x^{\alpha-1} e^{\beta x}}{\Gamma(\alpha)}\,.
$$
The gamma distribution has skewness $2/\sqrt{\alpha}$ and therefore cannot be approximated well by a normal distribution. We draw $2000$ samples from $p_\gamma(x,7.5,1)$ and train a single RTBM with three hidden nodes on the samples with the CMS-ES parameter bound set to $[-50,50]$. (Here and in the following examples, we take only samples with $|x| < 20$ into account, for numerical reasons of the theta function implementation.) The histogram of the training data together with the true underlying probability density and the resulting RTBM fit is shown in figure \ref{CauchyDistFig} (top left). Note that the RTBM was able to generate a good fit to the skewed distribution.

As another example, consider the Cauchy distribution with probability density
$$
p_C(x,x_0,\gamma) = \frac{\gamma}{\pi((x-x_0)^2+\gamma^2)}\,.
$$
In contrast to the normal distribution, this distribution possesses heavy tails and is therefore more difficult to model. We consider $p_C(x,0,1)$ and draw $1000$ samples as training input for a single RTBM with $N_{h}=3$ (with the parameter bound set to $40$). The resulting fit is shown in figure \ref{CauchyDistFig} (top middle) together with the sample and the true underlying density. Note that the heavy tails are clearly picked up by the RTBM fit.

In order to illustrate a mixture model, let us consider the mixture of Cauchy distributions given by
$$
m_C(v) =  0.6\, p_C(v,0,1)+0.1\, p_C(v,5,5) + 0.3\, p_C(v,10,1)\,.
$$
We set up an RTBM layer consisting of two RTBMs with $N_{h}=2$, \cf, figure \ref{RTBMmixturefig}, and train on a sample of $m(v)$ as above. The resulting fit is shown in figure \ref{CauchyDistFig} (top right). The two peaks are well captured by the fit. However, the tails of this particular fit turned out to be rather wrinkly, which is a characteristic of RTBM-based fits. This is clear from the discussions in section \ref{RTBMtheory}. Essentially, one can view $P(v)$ as a sort of Fourier approximation to other densities. We expect that by increasing the number of hidden nodes, or averaging over different runs, the quality of the fit can be further improved.

More one dimensional examples are shown on the bottom row of
figure \ref{CauchyDistFig}. For the next examples we always draw 5000
samples. On the bottom left plot we fit a single RTBM with three
hidden units (with the parameter bound set to 30) to a Gaussian mixture
defined as
$$
m_G(v)=0.6\,p_G(v,-5,3)+0.1\,p_G(v,2,2) + 0.3\,p_G(v,5,5)\,,
$$
where $p_G(v,\mu,\sigma)$ is the normal distribution. The RTBM achieves a good level of agreement with the underlying
distribution.

We also model the probability density function (for short pdf) defined in \cite{modha1994} (bottom
center) and the uniform distribution (right center) between $[-5,5]$
with a single RTBM with four and three hidden units, respectively (the
parameter bound is set to 30). We observe that in both cases, regions
where the underlying pdf is sharp and flat, which are in general difficult to model via neural networks, are reproduced reasonably well. As already mentioned above, the oscillatory effects are an artifact of the Fourier like approximation, and we expect that a larger number of hidden units can improve on this. (For this, a better implementation of the Riemann-Theta function would be desirable.)

\begin{figure}
  \begin{center}
  \includegraphics[scale=0.30]{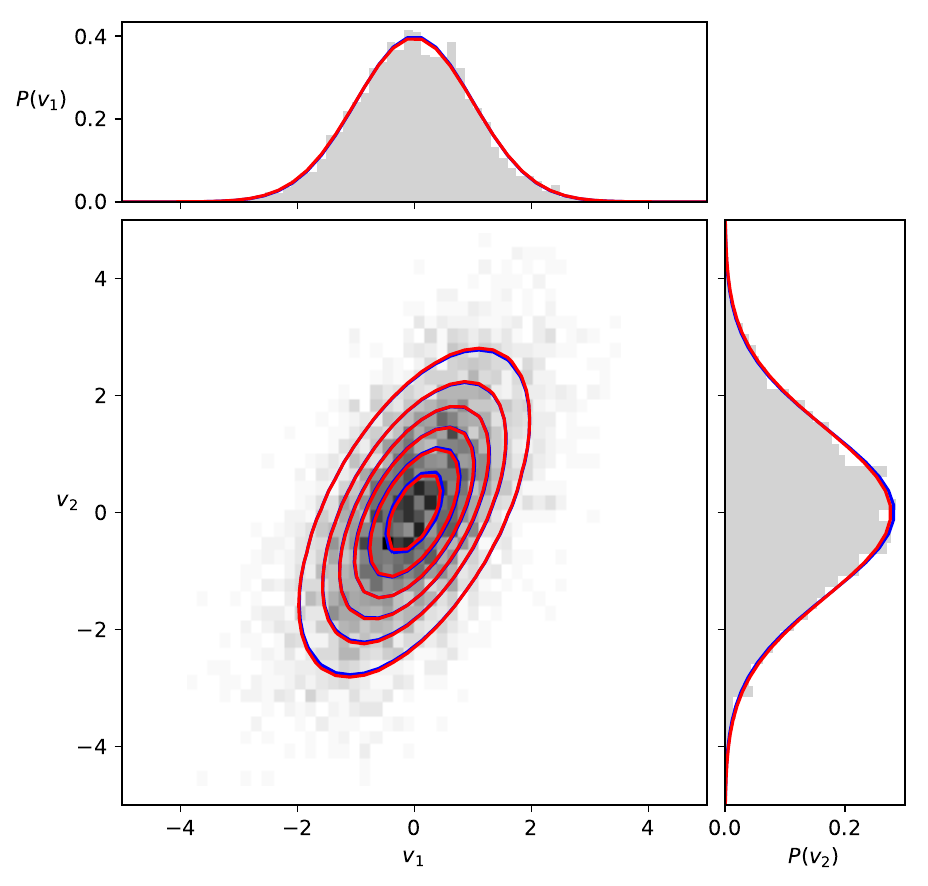}
  \includegraphics[scale=0.30]{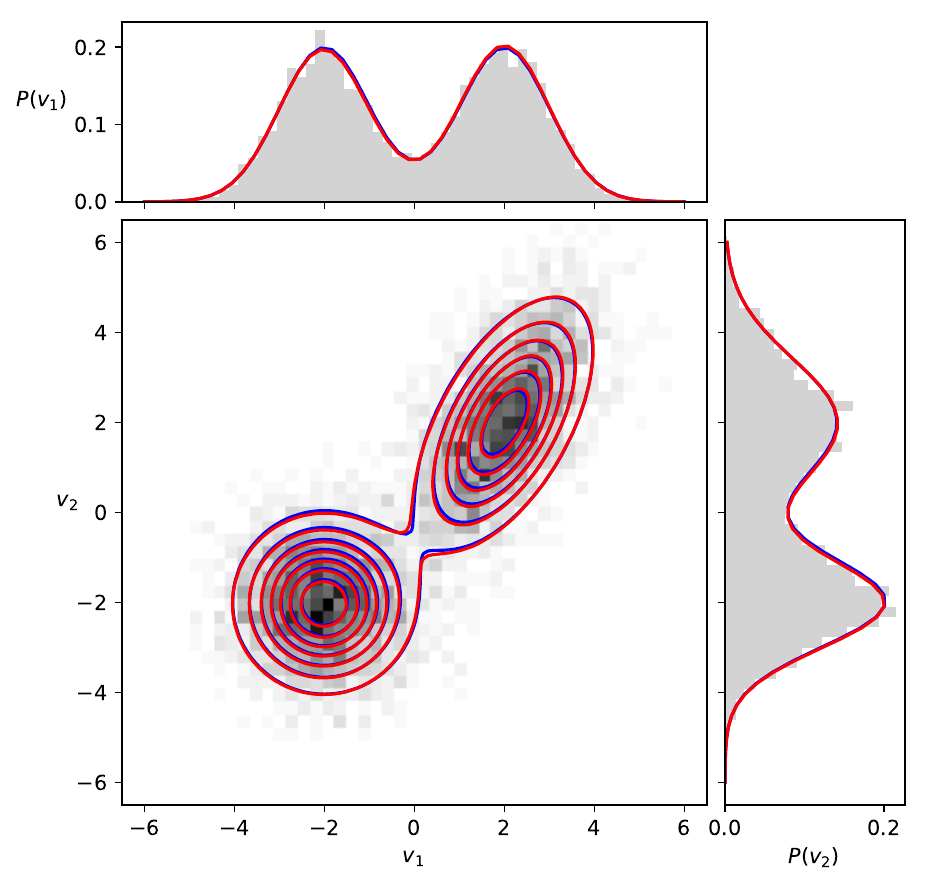}
  \includegraphics[scale=0.30]{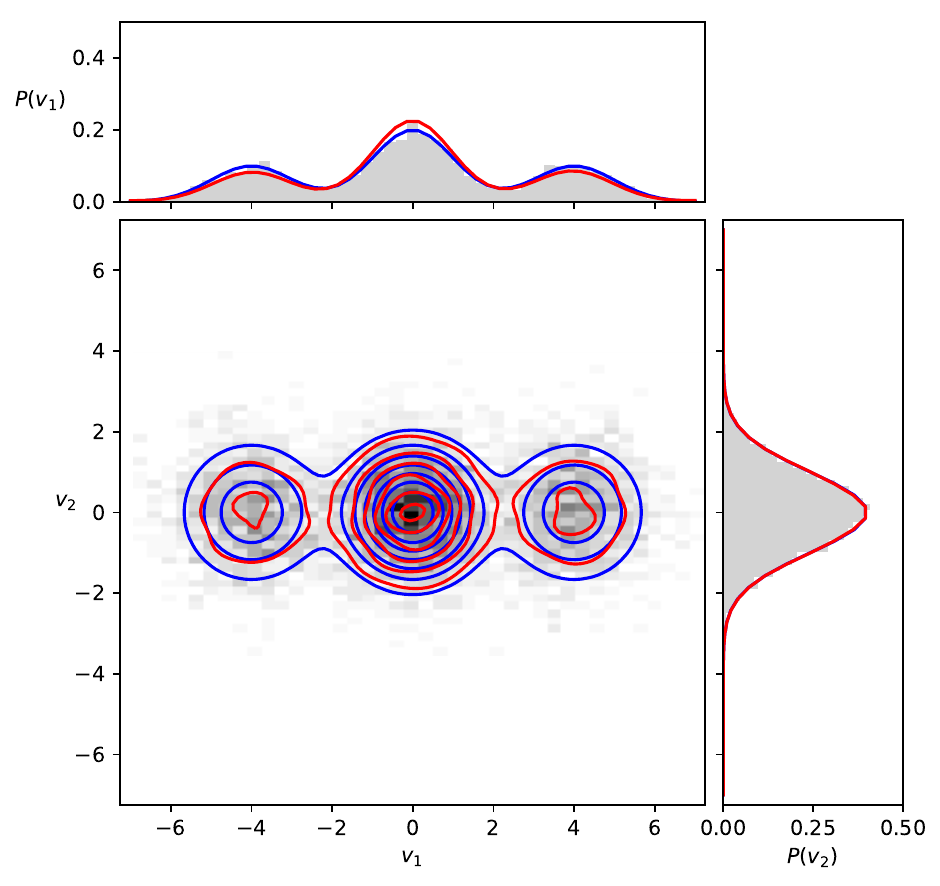}
  \caption{Left: Fit to a multivariate correlated Gaussian distribution via a single RTBM with $N_h=1$. Middle: Correlated multivariate Gaussian mixture fit using a layer of two RTBMs with $N_h=1$. Right: Gaussian mixture fit via a single RTBM with $N_h=2$. In all figures the contour plot of the trained model is shown together with its projections along the two axis. The blue line corresponds to the underlying true distribution, the red line to the fit and the histograms show the samples the models are trained on.}
  \label{2dDistFig}
  \end{center}
\end{figure}

In figure \ref{2dDistFig} we present three examples of two-dimensional
pdf determinations. In all plots we show the samples used by the fit
as a 2d projected histogram with a gray gradient color
map, while the contours of the underlying model and the trained RTBM fit are shown as blue and red (contour) lines. The side panels illustrate the projections along both axes.
On the left plot we model a correlated two-dimensional normal distribution centered at the origin which has covariance matrix $\Sigma$ with elements $\sigma_X^2=1$, $\sigma_Y^2=2$
and $\sigma_{X,Y}^2=0.8$, via a single RTBM with one hidden unit. 
The center plot shows a fit via an RTBM layer consisting of
two RTBMs with one hidden unit trained on samples of a Gaussian
mixture 
$$
m_G(v) = 0.5\, p_G(v,[2,2], \Sigma)+0.5\,p_G(v,[-2,-2], 1)\,.
$$
Finally, on the right plot we train a single RTBM
with two hidden units to fit the Gaussian mixture
$$
m_G(v) = 0.5\, p_G(v,[0,0],1)+0.25\,p_G(v,[4,0], 1)+0.25\,p_G(v,[-4,0],1)\,.
$$ 
For all examples we observe that the RTBM reproduces the underlying
distribution quite well. In order to quantify the quality of the RTBM
pdf estimate, in table \ref{MSETable} we compute the mean squared
error (MSE)
\begin{equation*}
  {\rm MSE^A_B} = \frac{1}{N_{\rm bins}} \sum_{i=1}^{N_{\rm bins}} (A_i - B_i)^2\,,
\end{equation*}
where the index $i$ refers to the bin index of the corresponding input
sampling histogram. Small values indicate good agreement between the
measured quantities.

\begin{table*}[t]
  \begin{center}
    \begin{tabular}{lcccc}
      \textbf{Distribution} & MSE$^{\rm pdf}_{\rm RTBM}$ & MSE$^{\rm pdf}_{\rm GMM}$ & MSE$^{\rm pdf}_{\rm GKDE}$ & MSE$^{\rm pdf}_{\rm CRBM}$ \tabularnewline
      \hline
      Gamma  & $ 6.8 \cdot 10^{-6}$ & $2.4 \cdot 10^{-5}$ [3] & $2.8 \cdot 10^{-5}$ [0.5] & $5.8 \cdot 10^{-3}$ \tabularnewline
      \hline
      Cauchy & $2.9 \cdot 10^{-5}$ & $8.1 \cdot 10^{-5}$ [10] & $1.5 \cdot 10^{-5}$ [0.4] & $4.8 \cdot 10^{-3}$ \tabularnewline
      \hline
      Cauchy mixture & $6.5 \cdot 10^{-5}$ & $7.6 \cdot 10^{-5}$ [9] & $3.3 \cdot 10^{-5}$ [0.5] & $2.9 \cdot 10^{-3}$ \tabularnewline
      \hline      
      Gaussian mixture & $1.9 \cdot 10^{-6}$ & $4.7 \cdot 10^{-6}$ [3] & $2.1 \cdot 10^{-6}$ [1.0] & $1.3 \cdot 10^{-3}$ \tabularnewline
      \hline
      Custom mixture & $6.7 \cdot 10^{-4}$ & $1.5 \cdot 10^{-3}$ [10] & $5.4 \cdot 10^{-4}$ [0.1] & $2.3 \cdot 10^{-2}$ \tabularnewline
      \hline
      Uniform & $8.2 \cdot 10^{-5}$ & $2.5 \cdot 10^{-4}$ [10] & $7.6 \cdot 10^{-5}$ [0.2] & $1.8 \cdot 10^{-3}$ \tabularnewline
      \hline
      \hline
      Corr. Gaussian & $2.7 \cdot 10^{-7}$ & $5.6 \cdot 10^{-7}$ [1] & $4.6 \cdot 10^{-6}$ [0.3] & $1.4 \cdot 10^{-6}$ \tabularnewline
      \hline
      Corr. Gaussian mixture & $2.2 \cdot 10^{-7}$ & $2.2 \cdot 10^{-7}$ [2] & $2.2 \cdot 10^{-6}$ [0.3] & $2.1 \cdot 10^{-5}$ \tabularnewline
      \hline
      Gaussian mixture & $5.8 \cdot 10^{-6}$ & $3.9 \cdot 10^{-7}$ [3] & $2.4 \cdot 10^{-6}$ [0.3] & $7.8 \cdot 10^{-6}$ \tabularnewline
      \hline
    \end{tabular}
    \caption{Distance estimators for the examples in figures \ref{CauchyDistFig} and \ref{2dDistFig}. The mean squared error (MSE) is taken between the fitting model and the underlying distribution (pdf). The numbers in the brackets correspond to the number of constituents of the gaussian mixture model (GMM), respectively to the bandwidth of the gaussian kernel density estimator (GKDE) model.}
    \label{MSETable}
  \end{center}
\end{table*}

We compare the MSE distances between the underlying distributions from
figures \ref{CauchyDistFig} and \ref{Examplesec} to the RTBM
predictions and three other common fitting models. Namely, a gaussian
mixture (GMM), a (gaussian) kernel density estimator (GKDE) and the continous
gaussian restricted Boltzmann machine (CRBM) with binary hidden
units, \cf, \cite{MWW2017}. The hyper-parameters of these models are
manually picked for the best fitting result. In particular, we use 20
hidden units for the CRBM with 10k training iterations making use of
the package \cite{PyDeep} (\texttt{GaussianBinaryVarianceRBM}). 

In general, we observe that the RTBM fit is superior to the CRBM and GMM fits, but can not consistently outperform the kernel based fits for the one dimensional examples. This is somewhat expected, as our examples are based on a large set of dense samples. It is interesting to note that in the last of the two dimensional examples we are able to model a multi-model distribution relatively well with a single RTBM, which would not be possible with a single gaussian.

\section{Theta neural networks}
\label{TNNsection}

\begin{figure}
\begin{center}
  \includegraphics[scale=0.3]{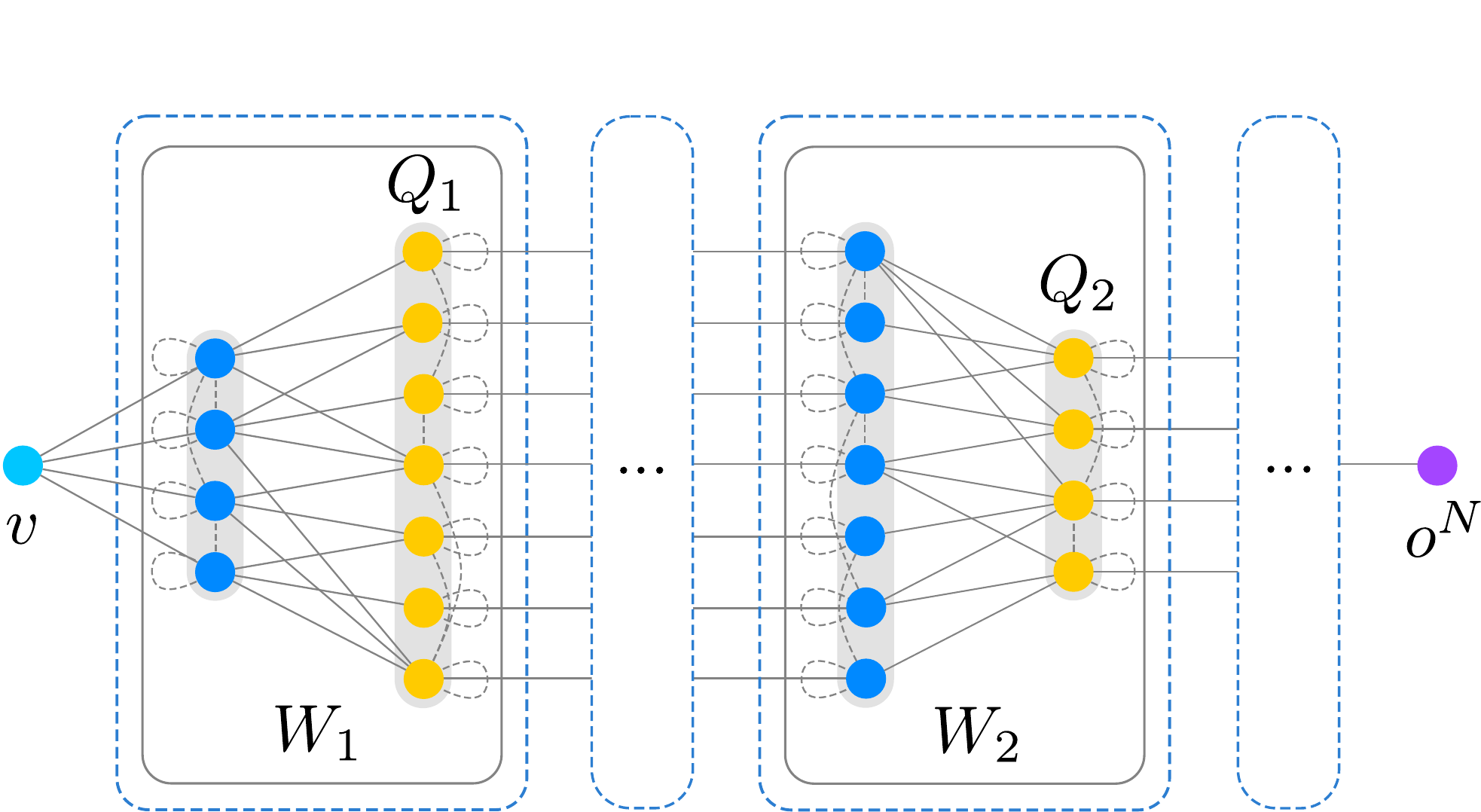}
  \caption{Illustration of a feedforward neural network with layers consisting of a RTBM. We take as $i$th output of a RTBM based layer the expectation $E(h_i|v)$. Note that the matrices $T$ do not enter the expectations. The RTBM layers can be arbitraly mixed and combined with other layers.}
  \label{TNNefig}
\end{center}
\end{figure}

The conditional expectation $E(h_i|v)$ can be used as an activation function in a feedforward neural network, replacing the usual non-linearities. In detail, for $Q$ of dimension $N_{h}\times N_{h}$ we can build a neural network layer consisting of $N_{h}$ nodes, with the output at the $i$th node given by $E(h_i|v)$. Here the inputs of the layer are given by the $v$ and we have the usual linear map $W$ occuring in \ref{EhvDef}. See also the illustration in figure \ref{TNNefig}.
The setup simplifies considerably if we restrict $Q$ to be diagonal due to the factorization property \ref{PhvFactorization}. In the diagonal case the activiation function at each node is independently given by the derivative of the logarithmic 3$rd$ Jacobi-Theta function, with its second parameter freely adjustable (\cf, \ref{EdDef}). Here, we will mainly consider this simplified setup due to its reduced computational complexity. Complex networks can be built by stacking such layers and inter-mixing them with ordinary neural network layers. We will refer to such networks which include RTBM-based layers as theta neural networks (TNN).

The gradients of the expectation unit can be calculated, and are given in appendix \ref{Egrads}. Hence the TNN can be trained as usual via gradient descent and backpropagation, in which case the additional parameters $Q$ can be treated similar to biases. However, as in the previous section it turns out that CMA-ES produces better results in particular examples, and therefore is currently our optimizer of choice.

\paragraph{Examples}

\begin{figure}
  \begin{center}
  \includegraphics[scale=0.33]{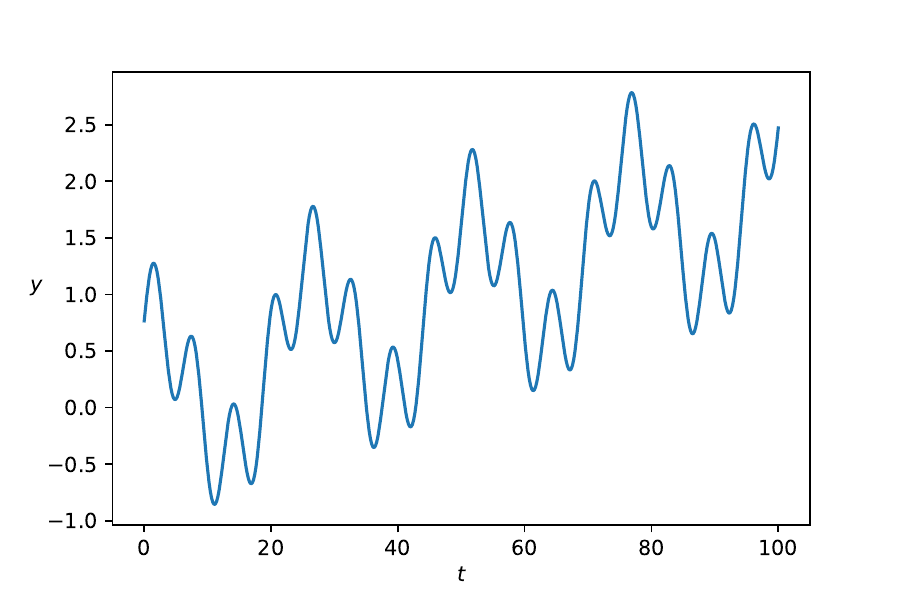}
  \includegraphics[scale=0.33]{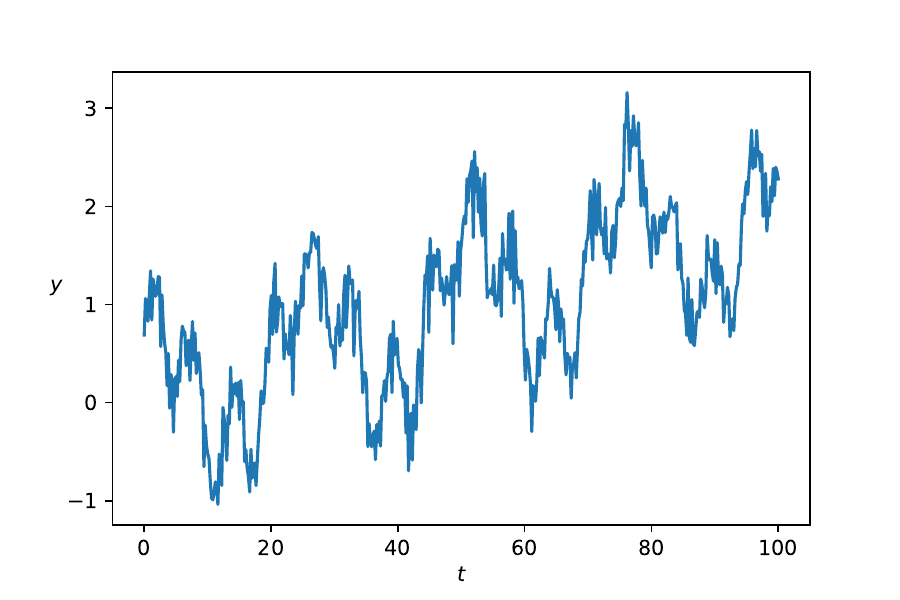}
  \includegraphics[scale=0.33]{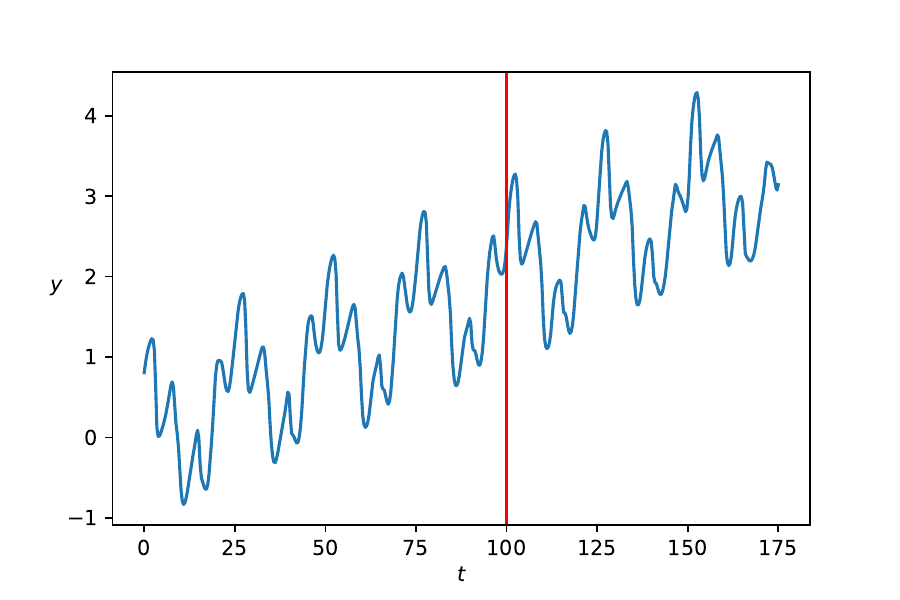}
  \caption{Left: The original signal given by a Sine-Cosine mixture with linear trend. Middle: With added gaussian noise. Right: The reconstructed signal from a sample of points and its extrapolation (right of red line)}
  \label{sincosmixfigs}
  \end{center}
\end{figure}

For illustration, let us consider a simple example. We want to learn the time-series
\beq\eqlabel{sincosmix}
y(t) = 0.02t+0.5\sin(t+0.1)+0.75\cos(0.25t-0.3)+\mathcal N(0,1)\,,
\eq
which is a sine-cosine mixture with linear trend and added Gaussian noise $\mathcal N(0,1)$. The signal with and without added noise is plotted in figure \ref{sincosmixfigs}. 
In order to learn the underlying signal, we set up a network with layer structure $1:3-3-2:1$, consisting of $E_{d}$ activation functions in phase I with 38 tunable parameters in total, making use of the library \cite{CK2017}. The network is trained on 500 pairs of $(t,y)$ values with $t\in [0,100]$, sampled from \req{sincosmix} via the CMA-ES optimizer with stopping criterium $10^{-4}$ using the MSE loss function. The learned signal and its extrapolation is shown in the right plot of figure \ref{sincosmixfigs}. The final MSE loss for the training and testing data sets are $3.6\cdot10^{-2}$ and $3.8\cdot10^{-2}$ respectively. The similarity between both values emphasizes that we were not only able to reconstruct the original signal from the noisy data on the training range, but also that the network learned the underlying systematics, as the extrapolation shows.

\begin{figure}
  \begin{center}
  \includegraphics[scale=0.5]{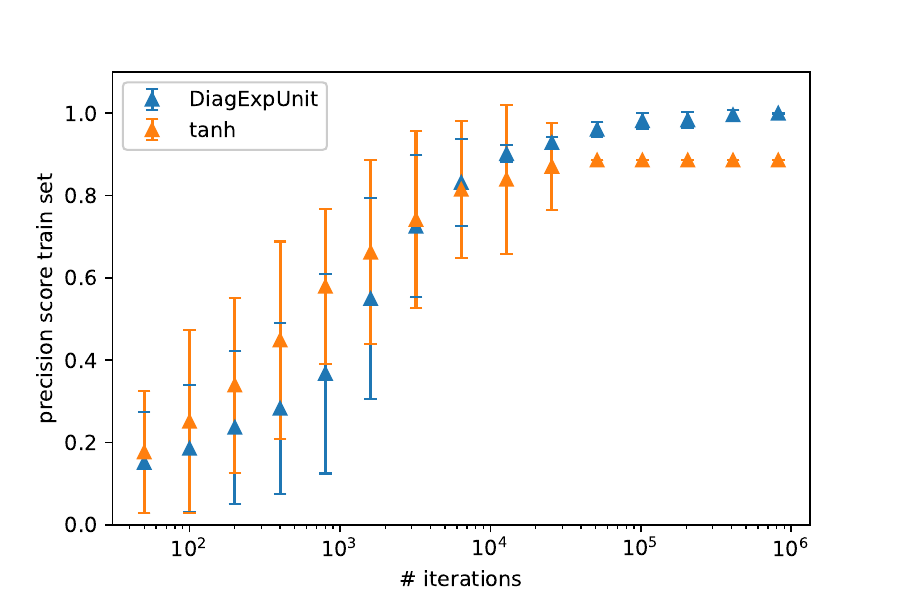}
  \includegraphics[scale=0.5]{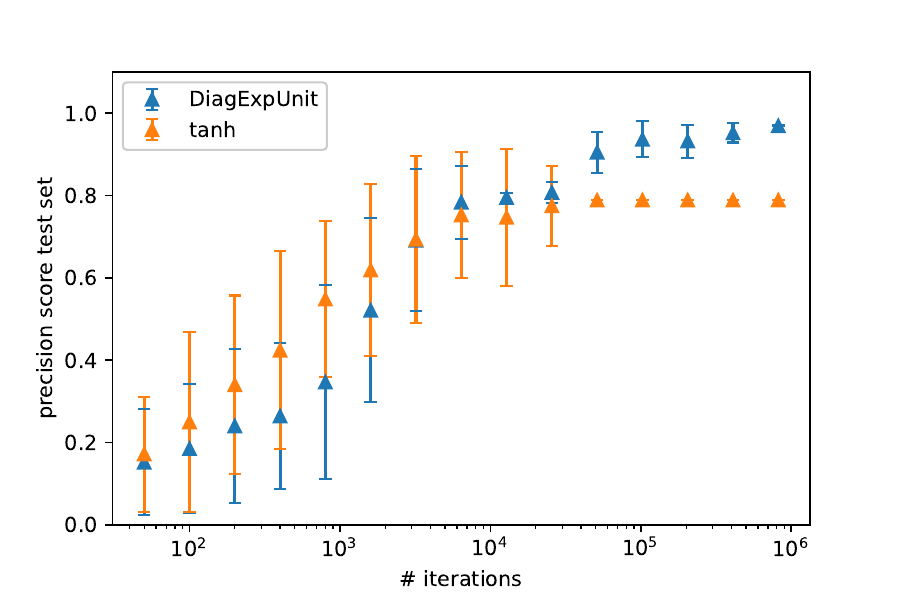}
  \caption{Left: Precision score achieved on the train set as a function of iterations. Right: Precision score achieved on the test set as a function of iterations. The score is ploted as the mean and standard deviation for 100 repetitions. The x-axis is plotted with a logarithmic scale. }
  \label{irisfigs}
  \end{center}
\end{figure}

As a classification example, let us consider the well known Iris data set \cite{iris1998}. This data set contains 150 instances from three different classes with four attributes. We reserve 40\% of the data as the test set. In order to investigate the modelling capacity of the $E_{d}$ activation functions, we set up two independent single-layer networks $4 : 3$ with the output unit activation functions in one network taken to be $E_{d}$, and in the other, $\tanh$. Both networks are trained via gradient descent and the adam optimizer for an increasing number of iterations in 100 independent repetitions. Note that for the initialization of the $Q$-parameters, we sample uniformly from the range $[2,18]$. The achieved precision scores are plotted in figure \ref{irisfigs}. 
We observe that the TNN-based classification converges more slowly, but ultimately achieves statistically significantly better classification rates than the network based on $\tanh$ units, on both the train and test data. 

\begin{figure}
  \begin{center}
  \includegraphics[scale=0.4]{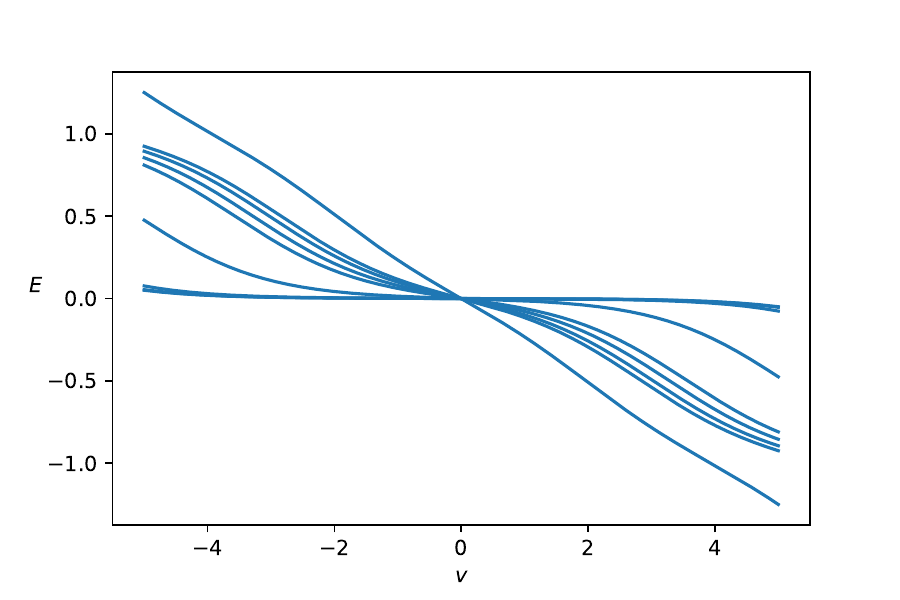}
  \includegraphics[scale=0.4]{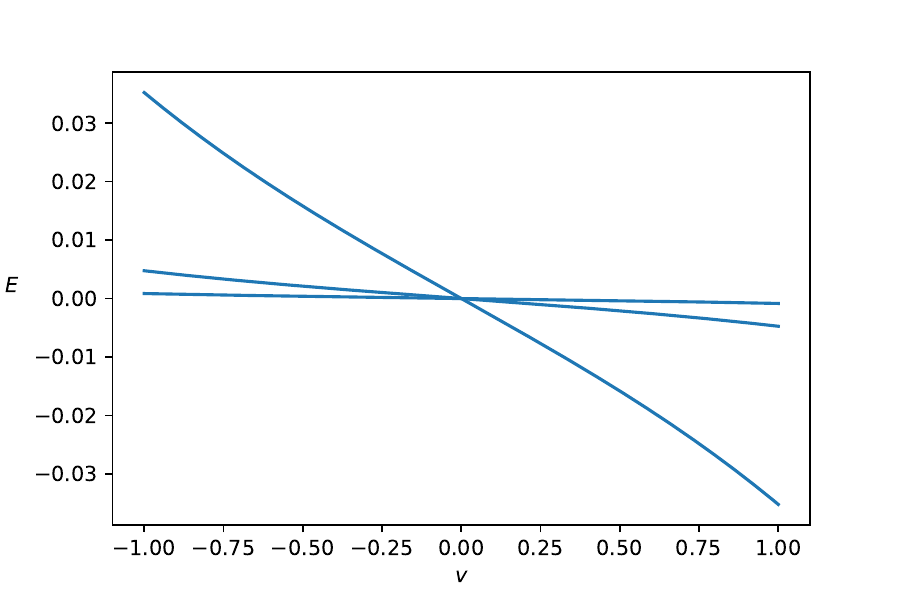}
  \caption{Left: The learned activation functions for the trending Sine-Cosine mixture. Right: Learned activation functions for the Iris dataset. Each line corresponds to $E_d(h|v)$ of a node.}
  \label{sincosmixact}
  \end{center}
\end{figure}

We plot the learned activation functions at each node for both toy examples discussed above in figure \ref{sincosmixact}. We observe that the TNNs learned a varity of activation functions, as theoretically expected.
We conclude that the additional learning of parameterization of the $E_{d}$ activation function (its shape) may have indeed benefits over a rigid activiation function like $\tanh$. Therefore TNN may be a promising direction for further research.

\section{RTBM classifier}
\label{RTBMclassSec}
\begin{figure}
\begin{center}
  \includegraphics[scale=0.25]{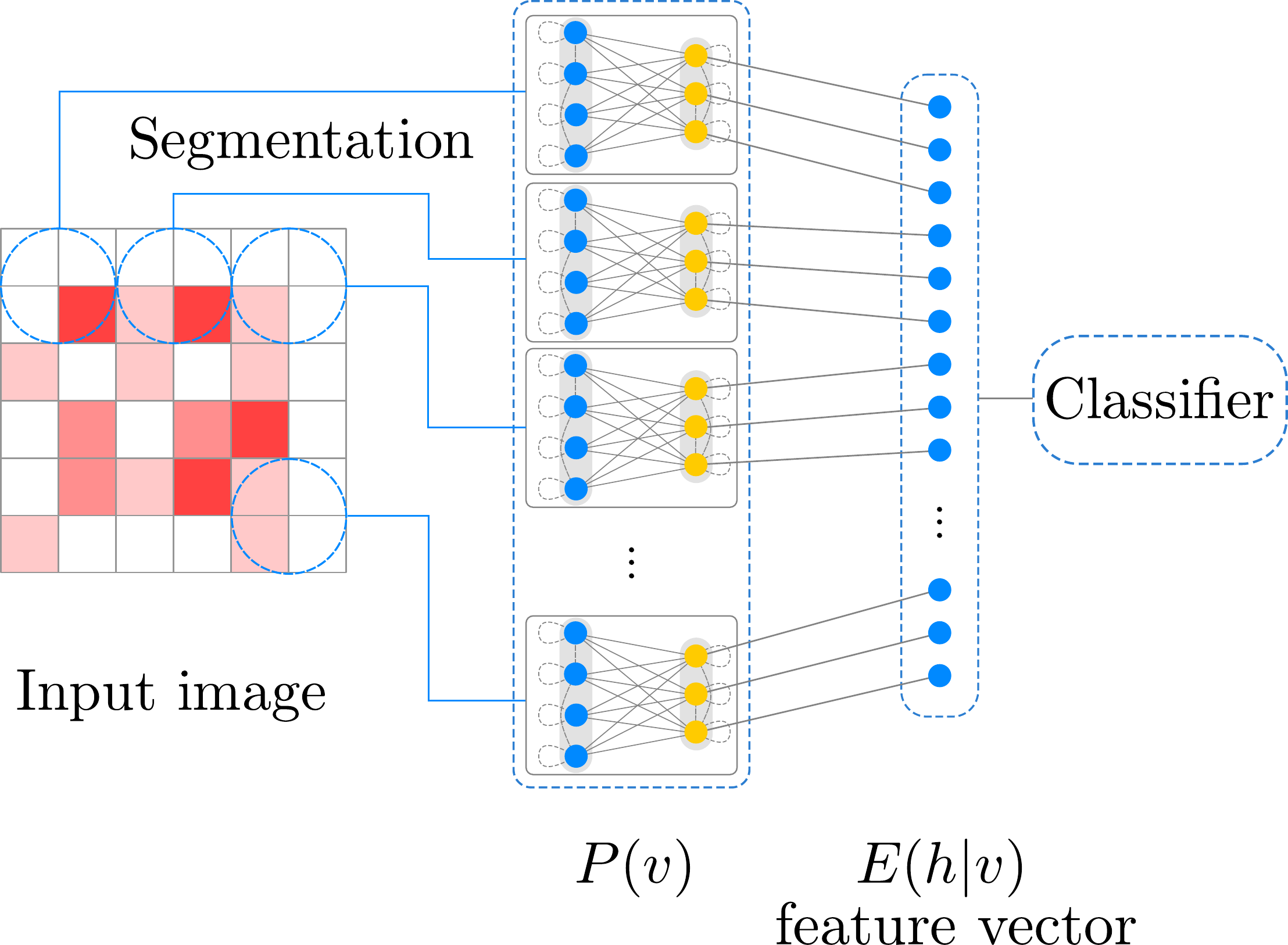}
\caption{Illustration
  of an RTBM classifier. An input image is segmented in blocks of four
  pixels. We model the probability density of each segment using
  a single RTBM. The resulting RTBM prediction in expectation mode is
  then fed into a standard classifier
  algorithm.} \label{RTBMclassification}
\end{center}
\end{figure}

The conditional expectation, which we already made use of in the previous section
to build TNNs, offers a further possibility to extend the applicability domain of RTBMs to
classification tasks. There are two possible ways to achieve data
classification through RTBMs. The first method consists of using TNNs, as in the previous section. In this case, the TNN classification requires the choice of
an appropriate cost function, usually the mean squared error, and an
adequate TNN architecture, which may contain extra layers which are not RTBM
based.
The second method follows \cite{K2009}. In the first step it segments
the input data into small patches. For each patch a single
RTBM model or RTBM mixture is used to generate the underlying
probability density of the input data. Then, for each input data we
collect the conditional expectation values for all hidden units of the
RTBM instances trained in probability mode. These expectation values are taken as a feature vector and are feed into a custom classifier. This method provides the
advantage of using the probability representation of RTBMs as an
autoencoder which preprocess the input data and simplifies the
classification task. Figure \ref{RTBMclassification} illustrates
schematically this method for an image example.

\paragraph{Example}
In order to show the potential capabilities of classification with
RTBMs, we have performed a short proof-of-concept test based on jet
substructure classification data from \cite{Baldi:2016fql}. The task
consists of discriminating between jets from single hadronic particles
and overlapping jets from pairs of collimated hadronic particles. For
this example we have selected 5000 images for training and 2500 for
testing. Each image is provided in a 32 pixel by 32 pixel format. As a reference
algorithm we use the logistic regression classifier which in the test
dataset scored a precision of 77\%.

The TNN regression obtained a precision score of 79\% using two RTBM
layers, the first with 1024 (32x32) visible units and three hidden
units and second layer with three input units and one hidden unit. The
precision values quoted for the reference and TNN regression have also
been tested on images resized using principal component analysis (PCA), showing negligible variations. The same network setup with however
hyperbolic non-linear activation funtions scored 55\% precision in our
tests.

The RTBM classifier obtained a precision score of 83\% with the
probability density determined by 50 RTBMs with two input and two hidden
units after resizing images to 10x10 pixels using PCA. The
classification is performed by logistic regression using as input
the expectation values from the 100 hidden units. We have verified
that the classification accuracy obtained with RTBMs in this example
is similar to results provided by simple neural networks (MLP) and
boosted decision trees. These results confirm that RTBM classifiers
could be interesting candidates for classification tasks.

\acknowledgments

We would like to thank G. van der Geer for valuable discussions and S. M. Tan for help in improving the manuscript. 
S.~C. is supported by the HICCUP ERC Consolidator grant (614577) and
by the European Research Council under the European Union's Horizon
2020 research and innovation Programme (grant agreement n$^{\circ}$
740006).

\appendix

\section{Continuous Boltzmann machine}
\label{CBM}
For completeness, we will briefly derive in this appendix the probability density $P(v)$ of the Boltzmann machine with continuous visible and hidden sector states. The setup is as in section \ref{RTBMtheory}.

The partition function in the continuous case reads
$$
Z = \int_{-\infty}^\infty[dv][dh]\, e^{-E(v,h)}\,,
$$
and can be calculated exactly making use of \req{gaussint}. We obtain
$$
Z = \frac{(2\pi)^{\frac{N_v+N_h}{2}}}{\sqrt{\det A}}\, e^{\frac{1}{2} B^t A^{-1} B}\,.
$$
The free energy now reads
$$
F = -\log\int_{-\infty}^{\infty}[dh]\, e^{-E(v,h)}\,,
$$
and evaluates to
$$
F=\frac{1}{2}v^t Tv+B_v^t v -\frac{N_h}{2}\log2\pi -\frac{1}{2}\log \det Q - \frac{1}{2} (v^t W +B^t_h)Q^{-1}(W^t v+B_h)\,.
$$
From the definition of the Boltzmann distribution \req{BoltzmannP} we obtain
$$
P(v)= \frac{e^{-\frac{1}{2}v^t(T-W Q^{-1}W^t)v +B_h^t Q^{-1} W^t v  - \frac{1}{2}B^tA^{-1}B +\frac{1}{2}B^t_h Q^{-1} B_h  }  }{(2\pi )^{\frac{N_v}{2}} \sqrt{\det((T-W Q^{-1}W^t)^{-1})}  } \,,
$$
where we made use of the determinantal formula for block matrices, giving $\det A = \det(Q)\det(T-W^t Q^{-1} W)$. The resulting probability density function is essentially a multi-variate Gaussian distribution with covariance matrix given by the inverse of the Schur complement
$$
(A/Q)^{-1} = (T-W Q^{-1}W^t)^{-1}\,.
$$
Hence, the Boltzmann machine with continuous visible and hidden sector is trivial.

\section{Gradients}
\label{gradientSection}
\subsection{$E(h_i|v)$}
\label{Egrads}
The gradients of the expectation unit \req{Eunit} can be easily calculated to be given by
\beq
\begin{split}
\frac{\partial E(h_i|v)}{\partial (B_h)_j} &= \kappa_{ji}\,,\\
\frac{\partial E(h_i|v)}{\partial W_{jk}} &= \kappa_{ki} v_j\,,\\
\frac{\partial E(h_i|v)}{\partial Q_{jk}} &= (1+\delta_{jk})^{-1}\rho_{jki}\,,
\end{split}
\eq
with
$$
\kappa_{ji} = -\frac{1}{(2\pi \ii)^2}\left(\frac{\nabla_j\nabla_i \theta_a}{\theta_a} -\frac{(\nabla_j \theta_a)(\nabla_i \theta_a)}{\theta_a^2}\right)\,,
$$
and
\beq
\rho_{jki} = \frac{1}{(2\pi\ii)^3}\left(\frac{\nabla_j\nabla_k\nabla_i \theta_a}{\theta_a} -\frac{(\nabla_j \nabla_k \theta_a)(\nabla_i \theta_a)}{\theta_a^2} \right)\,.
\eq
(We used the abbreviation $\theta_a = \tilde\theta(v^t W+B_h^t|Q)$.)
Note that in order to arrive at the derivative with respect to $Q$, we made use of the heat equation like relation 
\beq\eqlabel{heateqRelation}
\partial_{Q_{jk}}\tilde\theta(z|Q) = -\frac{1}{(2\pi\ii)^2}(1+\delta_{jk})^{-1}\nabla_j\nabla_k \tilde\theta(z|Q)\,,
\eq
which can be easily derived from the definition \req{RTdef}. 

\subsection{$P(v)$}
\label{PvGradients}
In order to calculate the gradients of the probability density \req{Pv} we make use of relation \req{heateqRelation} to infer that
\beq\eqlabel{Pgrads}
\begin{split}
\frac{\partial P(v)}{\partial(B_h)_i} &= \frac{P(v)}{2\pi\ii} \left( \frac{\nabla_i\theta_a}{\theta_a} - \frac{\nabla_i\theta_b}{\theta_b}  \right)\,,\\
\frac{\partial P(v)}{\partial(B_v)_i} &= P(v)\left( -v_i -(T^{-1}B_v)_i  + (T^{-1}W D_b)_i \right)\,,\\
\frac{\partial P(v)}{\partial Q_{ij}} &= -(1+\delta_{ij})^{-1}\frac{P(v)}{(2\pi\ii)^2}\left(\frac{\nabla_i\nabla_j\theta_a}{\theta_a}-\frac{\nabla_i\nabla_j\theta_b}{\theta_b} \right)\,,\\
\frac{\partial P(v)}{\partial W_{ij}} &= P(v)\left(v_i\frac{\nabla_j \theta_a}{\theta_a} + (B^t_v T^{-1})_i \frac{\nabla_j \theta_b}{\theta_b} -(H_b W^t T^{-1})_{ji} - (T^{-1}W H_b)_{ij} \right)\,,
\end{split}
\eq
with the normalized gradient vector and (rescaled) hessian matrix
$$
(D_b)_i:= \frac{1}{2\pi\ii}\frac{\nabla_i\theta_b}{\theta_b}\,,\,\,\,\,\,(H_b)_{ij}:=  \frac{(1+\delta_{ij})^{-1}}{(2\pi\ii)^2}\frac{\nabla_i\nabla_j\theta_b}{\theta_b}\,.
$$
(Note that we defined $\theta_b:=\tilde\theta\left(B_h^t - B_v^t T^{-1} W|Q-W^t T^{-1} W\right)$.)

The gradient with respect to $T$ requires that we restrict $T$ to be diagonal, such that $\det T =\prod_{i} T_{ii}$ and $(T^{-1})_{ii}=\frac{1}{T_{ii}}$. Under this restriction, we easily obtain
$$
\frac{\partial P(v)}{\partial T_{ii}} = P(v)\left(\frac{T_{ii}^{-1}+(B_v)_i^2 T^{-2}_{ii} - v_i^2}{2} -(B_v)_i T_{ii}^{-2}(WD_b)_i + T^{-2}_{ii} (W H_b W^t)_{ii} \right)\,.
$$

\section{Moments}
\label{MomentsSec}
We want to compute moments of the probability density $P(v)$. To this end note that we infer from \req{Pgrads}
\begin{equation}
	v_i P(v) = P(v)\left( -(T^{-1}B_v)_i + (T^{-1}W D_b)_i  \right) - \frac{\partial P(v)}{\partial(B_v)_i} \,.
\end{equation}
Using the normalization $\int [dv] \, P(v) = 1$, we immediately deduce that the first moments read
\begin{equation}
\begin{split}
	\langle v_i \rangle_P &\equiv \int [dv] \, v_i \,  P(v) = -(T^{-1}B_v)_i + (T^{-1}W D_b)_i  - \frac{\partial}{\partial(B_v)_i} \int [dv] P(v) \\
   &= -(T^{-1}B_v)_i + (T^{-1}W D_b)_i \,.
  \end{split}
\end{equation}

Similarly, we can compute the second moments
\begin{equation}
\begin{split}
	\langle v_i v_j \rangle_P & \equiv  \int [dv]\, v_i v_j \, P(v) \nonumber \\
	~ & = \langle v_i \rangle_P \langle v_j \rangle_P + T^{-1}_{ij} + \frac{(T^{-1} W)_{ik} (T^{-1} W)_{jl}}{(2\pi\ii)^2} \left(\frac{  \nabla_k \nabla_l \theta_b}{\theta_b} - \frac{\nabla_k  \theta_b \nabla_l  \theta_b}{\theta_b^2}\right)\,.
  \end{split}
\end{equation}
Higher order moments can be calculated analogously by taking more derivatives.

\end{document}